\documentclass[lettersize,journal]{IEEEtran}
\usepackage{amsmath,amssymb,amsfonts}
\usepackage{algorithmic}
\usepackage{algorithm}
\usepackage{array}
\usepackage[caption=false,font=normalsize,labelfont=sf,textfont=sf]{subfig}
\usepackage{textcomp}

\usepackage{fancyhdr}

\usepackage{booktabs}           
\usepackage{threeparttable}     
\usepackage{multirow}
\usepackage{makecell}           

\usepackage{graphicx}
\graphicspath{ {./figures/} }
\usepackage{subfig}

\usepackage{stfloats}
\usepackage{url}
\usepackage{verbatim}
\usepackage{cite}
\begin{document}

\title{Towards Top-Down Stereo Image Quality Assessment via Stereo Attention}

\author{Huilin Zhang, Sumei Li,~\IEEEmembership{Member,~IEEE,} Haoxiang Chang, Peiming Lin
\thanks{This work was supported by the National Natural Science Foundation of China under Grant 61971306. (Corresponding author: Sumei Li.)}
\thanks{Huilin Zhang, Sumei Li, Haoxiang Chang, and Peiming Lin are with the School of Electrical and Information Engineering, Tianjin University, Tianjin 300072, China (e-mail: hl\_zhang@tju.edu.cn; lisumei@tju.edu.cn; chang\_hx@tju.edu.cn; lpm2000@tju.edu.cn).}
}



\maketitle

\IEEEpeerreviewmaketitle
\fancypagestyle{firstpage}{
  \fancyhf{}
  \fancyhead[l]{\textit{This work has been submitted to the IEEE for possible publication. Copyright may be transferred without notice, after which this version may no longer be accessible.}}
}
\thispagestyle{firstpage}

\begin{abstract}
Stereo image quality assessment (SIQA) plays a crucial role in evaluating and improving the visual experience of 3D content. Existing visual properties-based methods for SIQA have achieved promising performance. However, these approaches ignore the top-down philosophy, leading to a lack of a comprehensive grasp of the human visual system (HVS) and SIQA. This paper presents a novel Stereo AttenTion Network (SATNet), which employs a top-down perspective to guide the quality assessment process. Specifically, our generalized Stereo AttenTion (SAT) structure adapts components and input/output for stereo scenarios. It leverages the fusion-generated attention map as a higher-level binocular modulator to influence two lower-level monocular features, allowing progressive recalibration of both throughout the pipeline. Additionally, we introduce an Energy Coefficient (EC) to flexibly tune the magnitude of binocular response, accounting for the fact that binocular responses in the primate primary visual cortex are less than the sum of monocular responses. To extract the most discriminative quality information from the summation and subtraction of the two branches of monocular features, we utilize a dual-pooling strategy that applies min-pooling and max-pooling operations to the respective branches. Experimental results highlight the superiority of our top-down method in advancing the state-of-the-art in the SIQA field. The code is available at \url{https://github.com/Fanning-Zhang/SATNet}.
\end{abstract}


\section{Introduction}
\label{sec:intro}
\IEEEPARstart{W}{e} have seen a lot of 3D-related multimedia products launched over the past decade. Toward creating a more immersive stereo world, stereo information processing technologies attract attention from both industry and academia. Since stereo images degrade inevitably during the acquisition, processing, and transmission, stereo image quality assessment (SIQA) has become the first imperative. In view of pristine images barely obtained in practical applications, no-reference SIQA (NR-SIQA) is exactly our focus.

Considering that the quality of a stereo image is evaluated by the human subjective perception in actual assessment, it is crucial to develop algorithms to mimic the human visual system (HVS) for effective NR-SIQA. Earlier methods have been explored in implementing HVS through traditional algorithms, and numerous literature~\cite{ZhouWujie-KNN,ZhouWujie-ELM,Fang-Access,LiuYun-Access,Messai-AdaBoost} concentrated especially on binocular properties (binocular fusion and rivalry). However, these works highly depend on hand-crafted features and are not robust and generalizable enough to represent inherent properties in stereo images. In the past few years, we have witnessed the prosperity of convolutional neural network (CNN) algorithms~\cite{Classification,segmentation,Detection} in the computer vision community, which can be attributed to CNN reflecting the philosophy of HVS to some extent, drawing inspiration from the signal-processing mechanism of neurons in the human brain.

\begin{figure}[t]
\centering
\includegraphics[width=\linewidth]{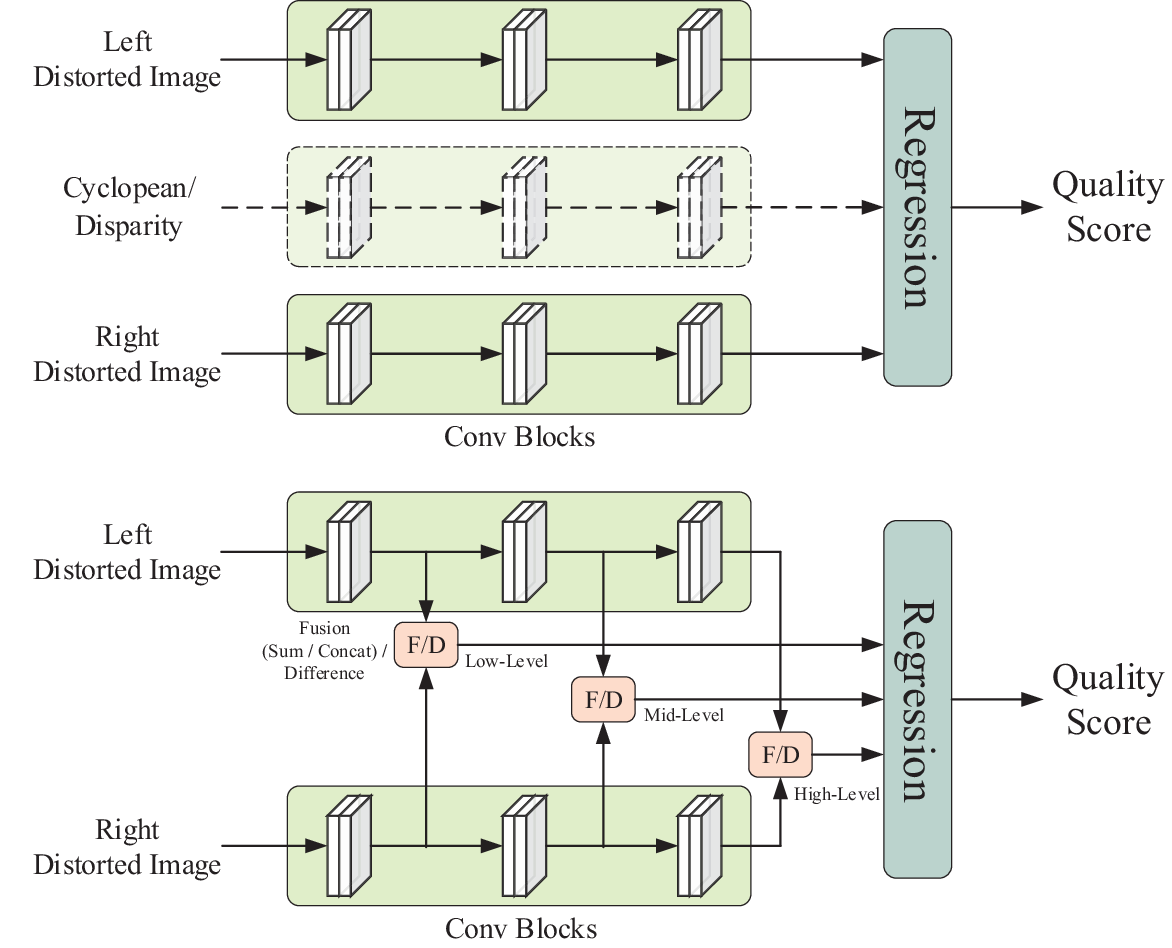}
\caption{Pipelines of two general bottom-up methods for NR-SIQA.}
\label{fig:Top-Down}
\end{figure}

Currently, CNN-based methods have been the mainstream approaches in the NR-SIQA field. For example, Zhang et al.~\cite{Zhang2016} first introduced CNN into the NR-SIQA field, proposing a three-column CNN model with the left view, right view, and their difference map as inputs. Fang et al.~\cite{Fang2019Siamese} proposed a Siamese network where two network branches share the same weights. The Siamese network outputs feature maps from two monocular branches, which are concatenated before regressing the quality score. Shi et al.~\cite{Shi2020PR} designed an RM-CNN3 model, employing a similar philosophy to~\cite{Zhang2016}, while their network focuses on multi-task SIQA. A one-column CNN model was constructed in~\cite{lsm-zhaoping}, where Cyclopean image patches were the only inputs. Although these works have achieved promising quality evaluation performance by utilizing the advantages of CNN architecture to mimic binocular properties, the interaction between monocular features or between binocular and monocular features (or both) is lacking or implicitly used during feature extraction as depicted in the upper part of Fig.~\ref{fig:Top-Down}, which is na\"{i}ve and inadequate. 

With diving deeper into HVS, more works have begun to simulate multi-level interaction. Zhou et al.~\cite{Zhouwei2019TIP} designed a dual-stream network with binocular interaction in multiple layers, and they concatenated fusion and difference maps that are regarded as interactive information. Bourbia et al.~\cite{Bourbia2021ICIP} employed and modified~\cite{Zhouwei2019TIP}'s model, leveraging naturalness analysis to develop a multi-task CNN for SIQA. Both~\cite{Zhouwei2019TIP} and~\cite{Bourbia2021ICIP} use multi-level difference and fusion features to facilitate the model's performance. A multi-level feature fusion network was proposed in~\cite{Yan2020}, where three types of features extracted from three different levels were concatenated and regressed as quality output. As illustrated in the lower part of Fig.~\ref{fig:Top-Down}, these networks, aiming to obtain multiple fusion or difference features (or both) as multi-level interactive information for further transmission, have gained better results. From the perspective of information interaction and reasoning mechanisms, such a design philosophy of existing SIQA models~\cite{Zhang2016,Fang2019Siamese,Shi2020PR,lsm-zhaoping,Zhouwei2019TIP,Bourbia2021ICIP,Yan2020} can be considered a bottom-up strategy. Nevertheless, this strategy has an inherent drawback: it struggles to adapt to quality-aware tasks that demand guidance from higher-level information down to lower-level one, i.e., these methods ignore the top-down philosophy, leading to a lack of a comprehensive understanding of HVS and SIQA.

In effect, in the field of cognitive psychology and neuroscience, there are two types of reasoning mechanisms termed bottom-up and top-down, respectively~\cite{1993topdown,2003topdown}. Bottom-up reasoning implies the flow of information from lower-level, such as primary visual stimuli, up to higher-level perceptual procedures. In contrast, top-down processing refers to the opposite direction, where the higher-level cognitive processes could be expectations and prior knowledge~\cite{shi2023topdown}. In the SIQA context, people's task is distinguishing the perceptual quality distortion and evaluating degraded images. This task is high-level because quality-related expectations, such as distorted types and degrees, will influence how we perceive and assess the quality of an image. In addition, literature~\cite{TPAMI2013} pointed out that higher-level feedback can modulate early sensory representations, enhancing or suppressing certain features based on contexts and expectations. Therefore, we intuitively came up with the idea to implement the top-down process by constructing a Stereo AttenTion (SAT) structure with tailored components and input/output for SIQA context, employing an attention map as the higher-level binocular guidance to monocular signals since it can reflect the expectations of human's most interests and relatively advanced compared to the input of the attention module. 

Undoubtedly, the attention mechanism is a burgeoning tool to simulate HVS in vision tasks. As plug-and-play modules, it is a common practice to insert self-attention modules sequentially with convolution blocks in a CNN-based architecture, which realizes the recalibration of the feature responses of a single input. When it comes to stereo pairs, we naturally raise a question: is the attention mechanism appropriate to cope with stereo vision like SIQA? Researchers have made some explorations. Zhou et al.~\cite{ZhouMingyue} used attention modules after the left and right fusion implementation, and Li et al.~\cite{lsm-zhaoping} inserted a dual attention block following the feature extraction from Cyclopean image patches. Both insert vanilla attention modules to explore inner relationships in combined features of two monocular views. Different from these approaches~\cite{ZhouMingyue,lsm-zhaoping} that are still bottom-up, our SAT structure adapts components and input/output for stereo scenarios, conducting top-down modulation from higher-level binocular features to monocular ones. We also realize consecutive recalibrations by stacking SAT blocks.

Furthermore, we introduce an Energy Coefficient (EC) into SAT structure to make it more sensible and well-performing in light of a fact that biological researchers have put forward, which discloses that binocular responses in primate primary visual cortex are less than the sum of monocular responses of two eyes~\cite{Mitchell-iScience}. Zhang et al.~\cite{PKU2022} also proved this fact by investigating neuronal responses in macaque V1 with two-photon calcium imaging. Whereas to the best of our knowledge, almost all previous works took all portions of two monocular responses into account, which is unreasonable to some extent. Thus, we introduce an EC into the SAT structure to adaptively learn suitable binocular response magnitudes. 

Moreover, for the SIQA task, good features should be able to capture quality properties such as noise and blur, which are usually the smaller values in the feature map. Nonetheless, only max-pooling is often employed for most prior SIQA studies, following feature encoding to reduce computational overhead and maintain local structure invariance. Toward adding quality-sensitive attributes to our model, we employ min-pooling while retaining max-pooling, named dual-pooling. We demonstrate the validity of our proposed dual-pooling strategy through empirical evaluation, showcasing its effectiveness in screening the most crucial structure and distortion information for quality regression. 

Overall, the contributions of this work are three-fold:
\begin{enumerate}
\item{From a top-down perspective, we propose a Stereo AttenTion Network (SATNet) for NR-SIQA, which realizes the guidance from higher-level binocular signals down to lower-level monocular signals, allowing progressive recalibration of both throughout the pipeline.}
\item{We design a generalized Stereo AttenTion (SAT) block, where components and input/output are adapted for stereo scenarios, and the attention map is leveraged as the higher-level binocular modulator of two lower-level monocular features, implementing the top-down philosophy. Moreover, an Energy Coefficient (EC) is introduced for binocular feature formation, reflecting the fact that binocular responses in the primate primary visual cortex are less than the sum of monocular responses.}
\item{To screen the most discriminative quality information from the summation and subtraction of the two branches of monocular features, we apply min-pooling and max-pooling to them, respectively, namely the dual-pooling strategy. Empirical evidence verifies the superiority of our strategy.}
\end{enumerate}

The remainder of this paper is organized as follows. In Section~\ref{sec:related}, a review of related work is given. In Section~\ref{sec:method}, we introduce the architecture and details of the proposed SATNet. Experimental results and analysis are given and discussed in Section~\ref{sec:experi}. In addition, Section~\ref{sec:experi} also gives the cross-dataset validation and ablation experiments. Finally, conclusions are drawn in Section~\ref{sec:conclu}.

\section{Related Work}
\label{sec:related}

\subsection{NR-SIQA Methods}
\label{sec:related_SIQA}
Eminent NR-SIQA methods have evolved through two phases. In the early stage, pioneers generally extracted hand-crafted features from stereo pairs and regressed them to quality scores by different machine learning tools. With the bloom of CNN, researchers have developed substantial CNN-based SIQA models focusing on integrating feature extraction and quality regression in an end-to-end way. Therefore, we review conventional and CNN-based NR-SIQA methods in this subsection.

For traditional methods, researchers manually extract features to represent HVS characteristics, then employ machine learning models, e.g., Support Vector Regression (SVR), K-Nearest Neighbors (KNN), Extreme Learning Machine (ELM), etc., to integrate artificial features into predicted quality scores. For example, Zhou and Yu developed a binocular response-based NR-SIQA method in~\cite{ZhouWujie-KNN}. Binocular responses of stereo pairs, including Binocular Energy Response (BER) and Binocular Rivalry Response (BRR), can be obtained. Furthermore, the final quality score can be derived from produced quality-predictive features by KNN. An NR-SIQA model based on the binocular combination and ELM was devised by Zhou et al.~\cite{ZhouWujie-ELM}. Two binocular combinations of stimuli are generated via different strategies, and then diverse binocular quality-aware features of the combinations are extracted by local binary pattern operators. Finally, an ELM is adopted to map obtained features to objective quality prediction. Fang et al.~\cite{Fang-Access} proposed a no-reference quality evaluator of stereo pairs, getting monocular and binocular visual properties from left and right images. Further, they employed SVR to gain the final results. Liu et al.~\cite{LiuYun-Access} designed a 3D image quality evaluator accounting for human monocular visual properties and binocular interactions. They modeled monocular and binocular information and utilized an SVR to combine these features and assess the quality. Messai et al.~\cite{Messai-AdaBoost} proposed an NR-SIQA network using a neural network Adaptive Boosting (AdaBoost). Cyclopean views of distorted left and right images can be acquired to extract features including gradient magnitude, relative gradient magnitude, and gradient orientation. Then, AdaBoost is employed to generate the final results.

These traditional works~\cite{ZhouWujie-KNN,ZhouWujie-ELM,Fang-Access,LiuYun-Access,Messai-AdaBoost} mainly include two steps: feature extraction and model regression. In general, explicitly modeling variable lower-lever visibility factors (e.g., color, luminance, intensity, structure, and depth) artificially and then regressing them to derive an overall quality estimation is an implementation of the bottom-up philosophy. Additionally, intrinsic disadvantages of hand-crafted features, such as complexity and imprecision, may lead to performance deterioration. Accordingly, conventional NR-SIQA methods can not achieve high consistency with subjective assessment.

In the past decade, CNN has significantly boosted multiple computer vision tasks, of course, NR-SIQA involved. As we mentioned in Section~\ref{sec:intro}, some preliminary CNN-based methods simulated binocular properties for SIQA. Zhang et al.~\cite{Zhang2016} took the first step in using CNN to estimate the quality of stereo pairs. Their pioneering work established a three-column CNN model and represented depth information with difference images. Extracted high-level features from the left, right, and difference maps are simply concatenated before regression. Fang et al.~\cite{Fang2019Siamese} proposed a Siamese network to learn high-level monocular semantic information. The outputs of two sharing-weights branches are concatenated to produce the final quality score. Shi et al.~\cite{Shi2020PR} built an RM-CNN3 model for multi-task SIQA, sharing a similar philosophy with~\cite{Zhang2016}. A one-column CNN model was constructed in~\cite{lsm-zhaoping}, where Cyclopean image patches were processed as binocular input. These primary methods~\cite{Zhang2016,Fang2019Siamese,Shi2020PR,lsm-zhaoping} cannot get the desirable evaluative capability due to the interactive information between monocular features or between binocular and monocular features (or both) lacking or implicitly used. 

Further investigations conducted fusion or difference (or both) operations to obtain the multi-level interactive information. Zhou et al.~\cite{Zhouwei2019TIP} devised a dual-stream network with multiple binocular interactions, and they utilized concatenation to fuse monocular features, fusion maps, and difference maps from different levels. Bourbia et al.~\cite{Bourbia2021ICIP} designed their multi-task CNN for SIQA likewise, multi-level difference and fusion features are exploited to facilitate the model's performance. A multi-level feature fusion network was proposed in~\cite{Yan2020}, where three types of features extracted from three different levels were concatenated and regressed as quality output. These works have strongly proved the significance of implementing multi-level information interaction in SIQA. 

In recent years, researchers have designed more complex networks to mimic human visual perception and pursue better results. Shen et al.~\cite{Shen2021} conducted a cross-fusion strategy after primary feature extraction to model the fusion of left and right views occurring in the V1 cortex. Aiming to handle asymmetric distortion assessment, Messai et al.~\cite{Messai2022ICIP} proposed a deep multi-score model for NR-SIQA. The multi-score CNN incorporates left, right, and stereo objective scores to extract the corresponding properties of each view. Sim et al.~\cite{Sim2022} pointed out that evaluating perceptual quality and understanding semantics cannot be thoroughly separated, so they proposed a blind SIQA method based on binocular semantic and quality channels. Considering both binocular interaction and fusion mechanisms of the HVS, a hierarchical StereoIF-Net~\cite{Si2022TIP} for NR-SIQA is proposed to simulate the whole quality perception of 3D visual signals in the human cortex. Benefiting from that, StereoIF-Net has made advancements in evaluation results. Although these CNN-based methods have reaped increasing performance through sophisticated model design, they are still bottom-up, i.e., visual signals are gradually integrated and processed from lower-level details to higher-level conception through the pipeline. Unlike them, our method models visual information interaction from a top-down perspective in an SAT-based fashion, which is more in line with the fact that higher-level visual signals exert influence on lower-level counterparts in the visual cognition process.

\subsection{Attention Mechanism}
The attention mechanism in computer vision aims to imitate the attention mechanism of HVS. To some extent, the attention mechanism is similar to the saliency theory, aiming to allocate more computational resources toward the most informative signal components. Attention has recently been extensively explored to facilitate various vision tasks. SENet~\cite{SE} exploits channel inter-dependencies to re-weight the output of convolution blocks. Both channel and spatial relationships are taken into account in CBAM~\cite{CBAM} to realize channel-wise and spatial-wise recalibration. In NLNet~\cite{non-local} and GCNet~\cite{GC}, self-attention is introduced to investigate long-range dependencies through non-local operations. SKNet~\cite{SK} utilizes self-attention for kernel size selection. 

However, few researchers pay attention to designing a particular attention module for stereo tasks. When it comes to the SIQA task, former researchers just inserted vanilla attention modules into their CNN architecture to boost performance. For instance, Zhou et al.~\cite{ZhouMingyue} used a SE block followed after the fusion operation in three scales of the receptive field to reflect the competitive characteristic of binocular fusion implicitly. Li et al.~\cite{lsm-zhaoping} inserted a dual attention block following the feature extraction from Cyclopean image patches, accounting for both channel and spatial attention. Both transferred existing attention blocks into the SIQA context without any modification, trying to explore inner relationships among input features. Compared to these approaches that are still bottom-up, our SAT structure conducts top-down modulation from higher-level binocular features to lower-level monocular ones by modifying components and input/output of original attention blocks. Specifically, we leverage a fusion-generated attention map to implement top-down perception for the SIQA scenario. More details can be found in Section~\ref{subsec:SAT}.

\begin{figure*}[t]
\centering
\includegraphics[width=\linewidth]{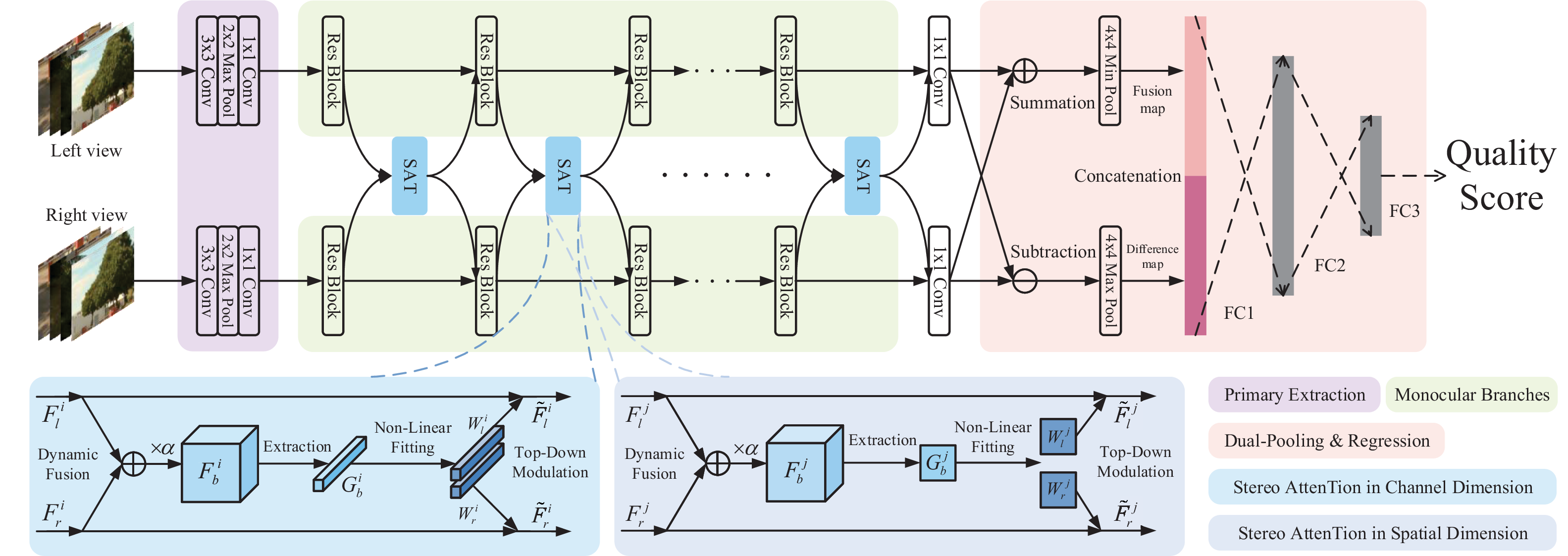}
\caption{The architecture of the proposed SATNet. Individual modules are termed in the corresponding color boxes in the lower right corner. In light that diverse attention modules focus on different attention dimensions, here we show two examples of SAT blocks in the channel and spatial dimensions.}
\label{fig:all}
\end{figure*}

\begin{table}[t]
    \centering
    \caption{The detailed configurations of proposed SATNet.}
    \label{tab:config}
    \resizebox{\linewidth}{!}{
        \begin{threeparttable}
        \begin{tabular}{*{5}{c}}
            \toprule      
            ~ & \textbf{Layer} & \makecell{\textbf{Configuration} \\ $S/P/BN/ReLU$} & \makecell{\textbf{Input Dimension} \\ $C\times H\times W$} & \makecell{\textbf{Output Dimension} \\ $C\times H\times W$} \\
            \midrule
            \multirow{3}{*}{\makecell{\textbf{Primary} \\ \textbf{Extraction}}} & $3\times3$ Conv & $1/1/\checkmark/\checkmark$ & $3\times40\times40$ & $64\times40\times40$ \\
            ~ & $2\times2$ Max Pool & $2/-/-/-$ & $64\times40\times40$ & $64\times20\times20$ \\
            ~ & $1\times1$ Conv & $-/-/-/-$ & $64\times20\times20$ & $64\times20\times20$ \\
            \midrule
            \multirow{2}{*}{\makecell{\textbf{Basic Block in} \\ \textbf{Monocular Branches}}} & $3\times3$ Conv & $1/1/\checkmark/\checkmark$ & $64\times20\times20$ & $64\times20\times20$ \\
            ~ & $3\times3$ Conv & $1/1/\checkmark/\checkmark$ & $64\times20\times20$ & $64\times20\times20$ \\
            \midrule
            \multirow{3}{*}{\makecell{\textbf{Bottleneck Block} \\ \textbf{in} \\ \textbf{Monocular Branches}}} & $1\times1$ Conv & $-/-/\checkmark/\checkmark$ & $64\times20\times20$ & $64\times20\times20$ \\
            ~ & $3\times3$ Conv & $1/1/\checkmark/\checkmark$ & $64\times20\times20$ & $64\times20\times20$ \\
            ~ & $1\times1$ Conv & $-/-/\checkmark/\checkmark$ & $64\times20\times20$ & $64\times20\times20$ \\
            \midrule
            \textbf{Stereo AttenTion} & ~ & ~ & $64\times20\times20$ & $64\times20\times20$ \\
            \midrule
            ~ & $1\times1$ Conv & $-/-/-/-$ & $64\times20\times20$ & $64\times20\times20$ \\
            \midrule
            \multirow{5}{*}{\makecell{\textbf{Dual-Pooling} \\ \textbf{\&} \\ \textbf{Regression}}} & $4\times4$ Min Pool & $4/-/-/-$ & $64\times20\times20$ & $64\times5\times5$ \\
            ~ & $4\times4$ Max Pool & $4/-/-/-$ & $64\times20\times20$ & $64\times5\times5$ \\
            ~ & FC1 & $-/-/-/\checkmark$ & $3200$ & $1600$ \\
            ~ & FC2 & $-/-/-/\checkmark$ & $1600$ & $800$ \\
            ~ & FC3 & $-/-/-/-$ & $800$ & $1$ \\
            \bottomrule      
        \end{tabular}
    \begin{tablenotes}
    \footnotesize
    \item[$S$] Stride
    \item[$P$] Padding
    \item[$BN$] Batch Normalization
    \item[$ReLU$] ReLU activation function
    \item[$C\times H\times W$] Channel$\times$Height$\times$Width
    \item[$-$] indicates no corresponding parameter
    \item[$\checkmark$] indicates that BN/ReLU is equipped in the corresponding layer
    \end{tablenotes}
    \end{threeparttable}
    }
\end{table}

\section{Proposed Method}
\label{sec:method}

\subsection{Network Architecture}
Our proposed SATNet is an end-to-end CNN. It takes left and right image patches sized $40\times40$ as dual-channel inputs and predicts a quality score as the final output. Fig.~\ref{fig:all} shows the pipeline. SATNet consists of four main parts: Primary Extraction, two monocular branches, SAT blocks, and Dual-Pooling \& Regression, and their names are labeled in the corresponding color boxes in the lower right corner of Fig.~\ref{fig:all}.

First, stereo image patches are encoded in the Primary Extraction phase, in which two independent convolution layers sized $3\times3$ and $1\times1$ extract monocular features in left and right branches. Between two convolutions, a $2\times2$ max pooling layer exists, downsampling the feature maps to $20\times20$ pixels so subsequent layers could process signals more efficiently. Then, we employ ResNet as the backbone of two monocular branches, leveraging its strong performance and efficiency in extracting features. The central part of SATNet is several chain-stacked res blocks, with each pair of res blocks followed by an SAT block. Given that our SAT blocks are refined from existing general attention modules and diverse attention modules focus on different attention dimensions, we show two examples of SAT blocks in the channel and spatial dimensions in the lower part of Fig.~\ref{fig:all}, respectively. After that, we use two convolution layers with $1\times1$ kernel size again to decode the outputs of two monocular branches separately. Next, as Fig.~\ref{fig:all} depicts, a Dual-Pooling \& Regression module is constructed. We perform summation and subtraction operations on two monocular features to generate fusion and difference maps, respectively. It is worth noting that all operations prior do not change the resolution of feature maps. Afterward, obtained fusion maps are $4\times4$ min-pooled, while difference maps are $4\times4$ max-pooled. Finally, fusion and difference maps sized $5\times5$ are flattened and concatenated, then three fully connected (FC) layers regress them to a quality score. During the inference stage, the quality score of a complete stereo pair is calculated by averaging all scores of patches from it. The details of two monocular branches, SAT blocks, and Dual-Pooling \& Regression are elaborated in the following subsections.

\subsection{Monocular Branches}
\label{subsec:MonoBran}
\begin{figure}[t]
\centering
\resizebox{\linewidth}{!}{
\subfloat[BasicBlock (depth=2)]{\label{subfig:BasicBlock}
\begin{minipage}[t]{0.5\linewidth}
\centering
\includegraphics[width=0.7\linewidth]{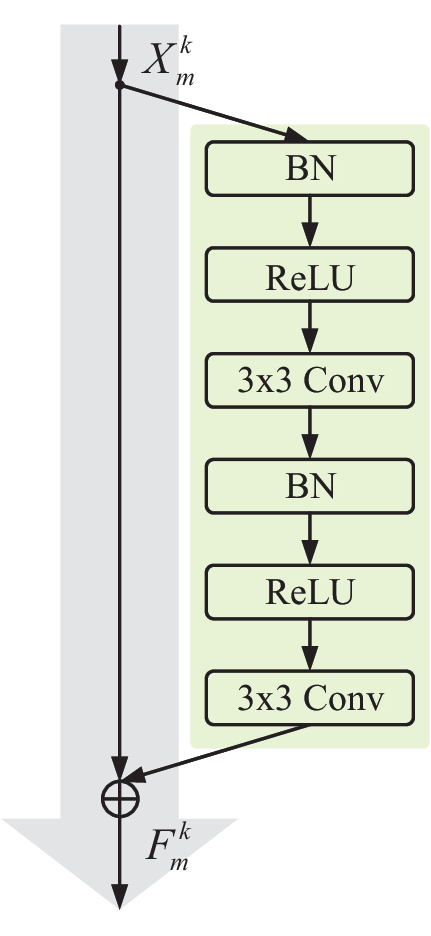}
\end{minipage}
}
\subfloat[Bottleneck (depth=3)]{\label{subfig:Bottleneck}
\begin{minipage}[t]{0.5\linewidth}
\centering
\includegraphics[width=0.7\linewidth]{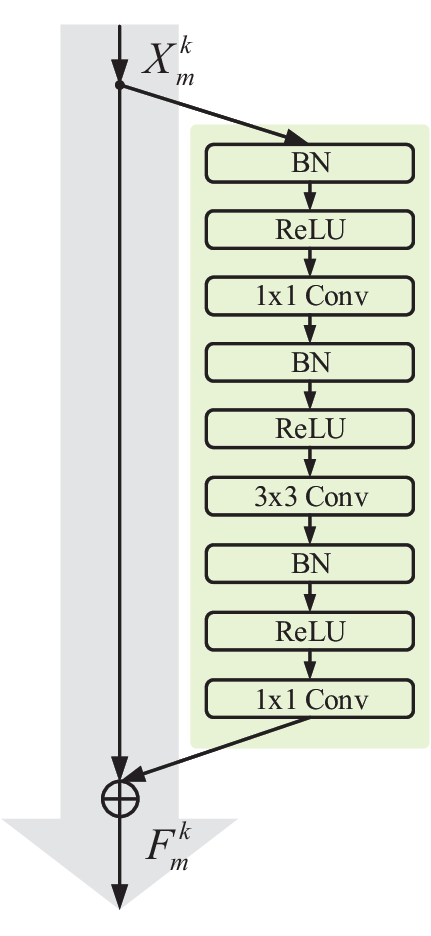}
\end{minipage}
}
}
\caption{Two variants of Res Block with full pre-activation.}
\label{fig:ResBlock}
\end{figure}

He et al.~\cite{resblock} proposed the renowned ResNet of five scales with corresponding residual blocks. Specifically, shallow models (ResNet-18/34) are composed of basic blocks with two convolutional layers with $3\times3$ kernel size, and deeper networks (ResNet-50/101/152) are designed with bottleneck units considering the imperative of reducing computational overhead in the meantime of remaining model's depth. The bottleneck block includes two convolutional layers sized $1\times1$ and a convolutional layer with $3\times3$ kernel size in the middle. To enhance the learning ability of the feature representation and avoid the gradient vanishing problem, we adopt residual blocks to establish a dual-stream ResNet as the backbone of monocular branches. Different from the original residual blocks~\cite{resblock}, we employ the better residual blocks with full pre-activation to boost performance, which benefits from enhanced identity mappings~\cite{resblock-v2}. Two blocks are illustrated in Fig.~\ref{fig:ResBlock}. 

In our design, the left and right res blocks of the same level are followed by an SAT block. So we construct our model with different scales with a range of SAT block numbers ($K$) referring to~\cite{resblock}. Here we assign $K=3,7,14,15$, respectively, corresponding to four scales of models (depth $=11,19,33,50$, respectively). We employ basic blocks to compose shallow models (in this case, $K=3,7,14$, corresponding to depth $=11,19,33$) while building the deep model via bottleneck blocks ($K=15$, depth $=50$), and model scales' configurations can be found in Table~\ref{tab:abla-scale}. The detailed configurations of SATNet are listed in Table~\ref{tab:config}. Let $X_m^k\in \mathbb{R}^{C\times H\times W}$ ($m\in \{l,r\},\ k=1,2,\ldots,K$) be the input feature of the $k$-th res block, the output $F_m^k\in \mathbb{R}^{C\times H\times W}$ can be described as 
\begin{equation}
    F_m^k=f_m(X_m^k,\omega_m^f),
\end{equation}
where $f_m(\cdot,\omega_m^f)$ is the abstract res block function for monocular branch $m\in \{l,r\}$. The left and right branches do not share parameters, and $\omega_m^f$ represents corresponding parameters.

\subsection{Stereo AttenTion (SAT)}
\label{subsec:SAT}
\begin{figure}[t]
\centering
\resizebox{\linewidth}{!}{
\subfloat[SAT]{\label{subfig:attn}
\begin{minipage}[t]{0.25\linewidth}
\includegraphics[width=\linewidth]{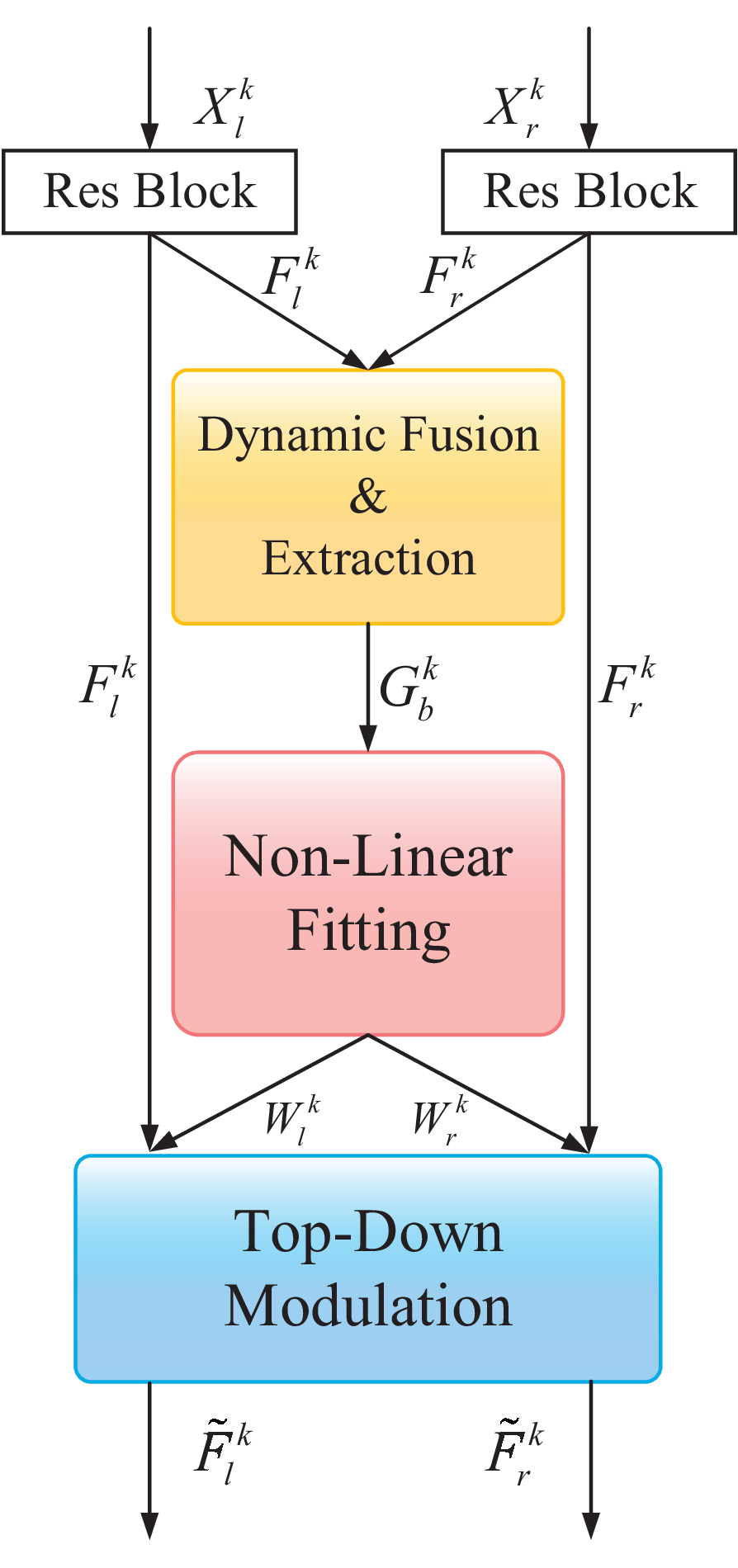}
\end{minipage}
}
\subfloat[SAT-SE]{\label{subfig:attn-SE}
\begin{minipage}[t]{0.25\linewidth}
\centering
\includegraphics[width=\linewidth]{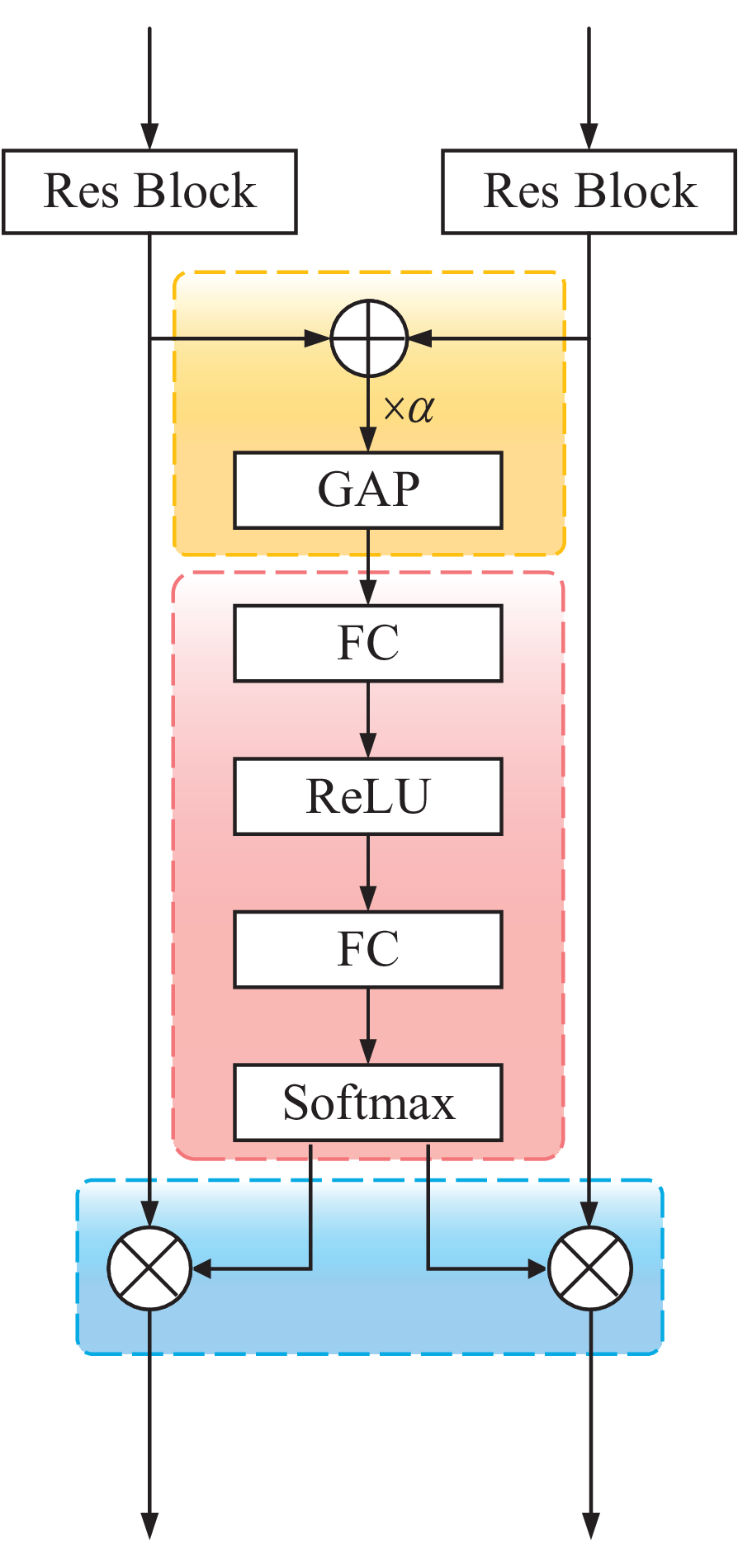}
\end{minipage}
}
\subfloat[SAT-CBAM]{\label{subfig:attn-CBAM}
\begin{minipage}[t]{0.25\linewidth}
\includegraphics[width=\linewidth]{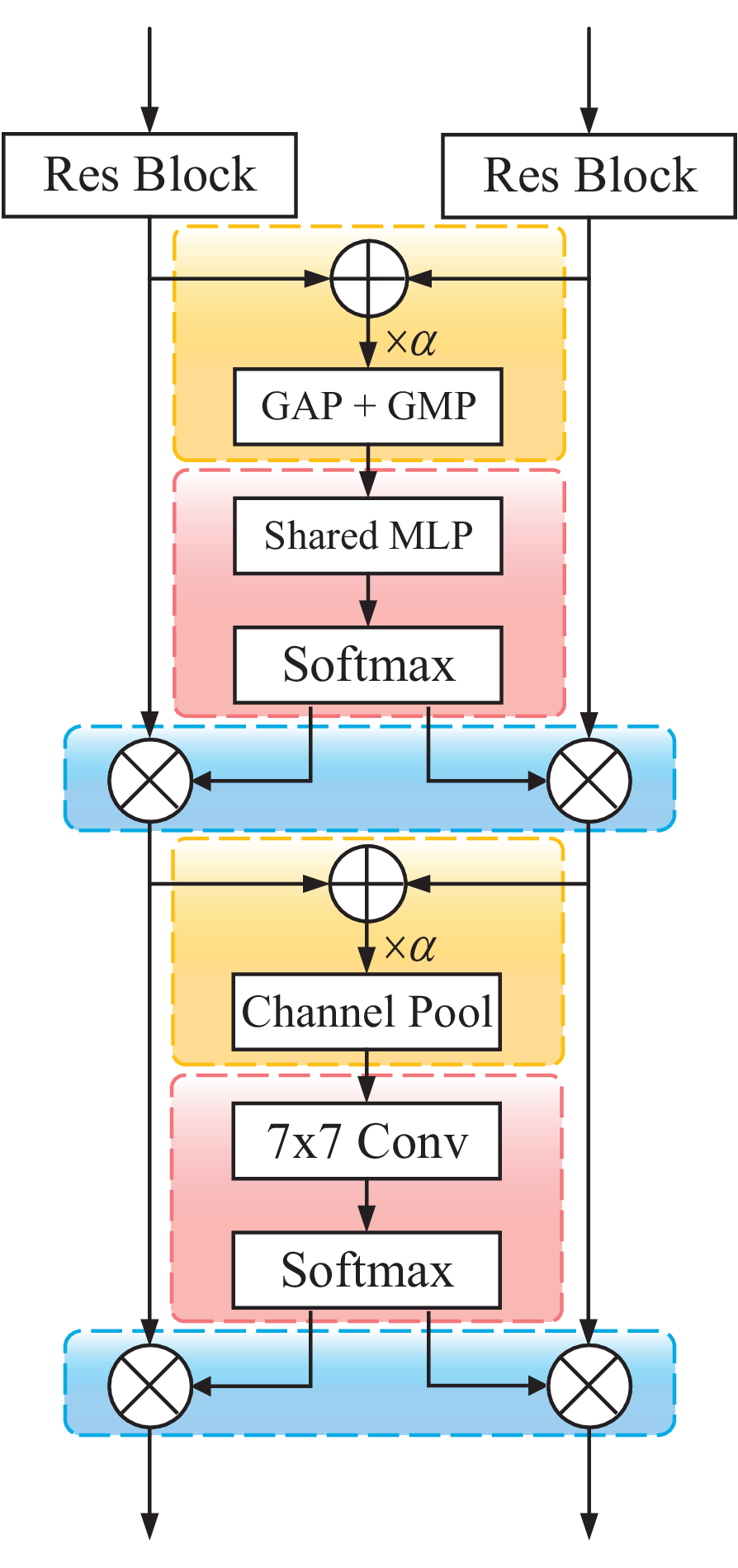}
\end{minipage}
}
\subfloat[SAT-GC]{\label{subfig:attn-GC}
\begin{minipage}[t]{0.25\linewidth}
\includegraphics[width=\linewidth]{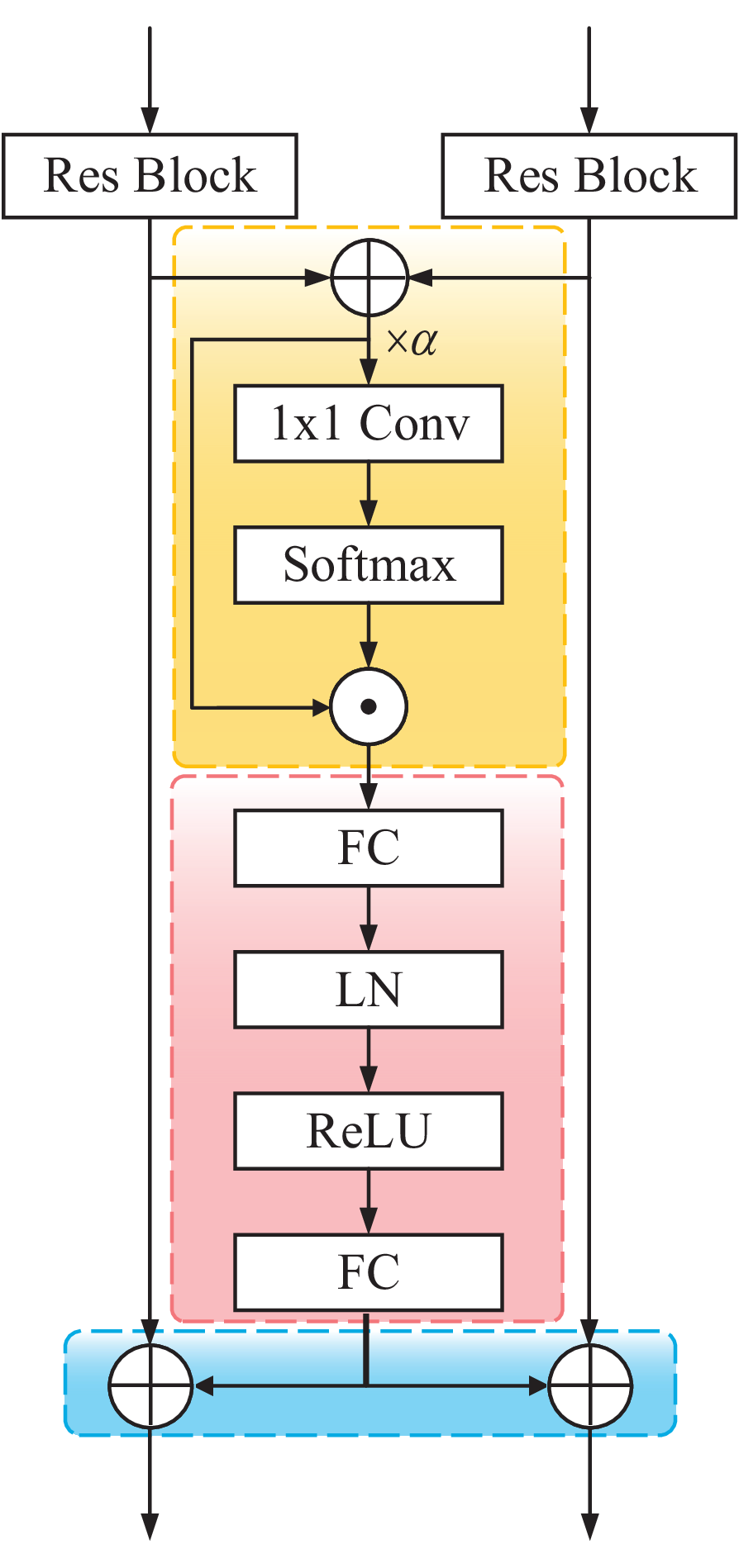}
\end{minipage}
}
}
\caption{We model a Stereo AttenTion (SAT) block with three components: Dynamic Fusion \& Extraction, Non-Linear Fitting, and Top-Down Modulation. Here we show three instances of this structure: \protect\subref{subfig:attn-SE} SAT-SE, \protect\subref{subfig:attn-CBAM} SAT-CBAM, and \protect\subref{subfig:attn-GC} SAT-GC. ``$\oplus$'' denotes element-wise summation, ``$\otimes$'' represents element-wise multiplication, and ``$\odot$'' performs matrix multiplication.}
\label{fig:attn-all}
\end{figure}

SAT structure is established to conduct top-down philosophy for SIQA, and it is inborn to model binocular responses, benefiting from information aggregation and non-linear abilities of attention modules. Inspired by~\cite{non-local,GC}, in which general attention module framework and components have been studied, we formulate SAT structure for stereo scenarios in three components: \textbf{Dynamic Fusion \& Extraction}, \textbf{Non-Linear Fitting}, and \textbf{Top-Down Modulation}. These components are generic and not confined to a particular attention block~\cite{ConnecAttn}. Considering the diverse attention dimensions targeted by various attention modules, we present two instances of SAT blocks, one in the channel dimension ($k=i$) and the other in the spatial dimension ($k=j$), depicted in the lower part of Fig.~\ref{fig:all}. For building SATNet for the NR-SIQA task, we also exemplify three classic attention blocks (SE~\cite{SE}, CBAM~\cite{CBAM}, GC~\cite{GC}) and their modeling by using our SAT structure, where the input/output is adapted for stereo contexts as well. Details are shown in Fig.~\ref{fig:attn-all}. 

\subsubsection{\textbf{Dynamic Fusion \& Extraction}}
It is designed to achieve two functions: Dynamically integrating features of two monocular branches and further extracting critical information. Here we introduce an EC ($\alpha$) to adaptively control the magnitude of binocular fusion features $F_b^k$ based on different inputs. Learnable $\alpha$ is constrained into $(0, 1)$ and multiplied by the summation of inputs $F_l^k$ and $F_r^k$ of an SAT block. We can write as
\begin{equation}
F_b^k=\alpha\times(F_l^k+F_r^k),
\end{equation}
where ``$\times$'' denotes element-wise multiplication and ``$+$'' indicates element-wise summation. $\alpha$ varies according to different factors, e.g., different SAT blocks, different phases ($k$), and different input features ($F_l^k$ and $F_r^k$). Thus, during the inference stage, each response magnitude of the binocular fusion feature is regulated accurately by the corresponding EC, which is more consistent with the actual perception process and beneficial to boost network performance. 

We define the feature extraction function after dynamic fusion as $g(\cdot,\omega^g)$, where $\omega^g$ is the parameter of the extractor $g$. Since pooling operation is often used for integration (i.e., SE, CBAM), $g$ is usually a parameter-free operation ($\omega^g$ is not needed) and can be simplified as $g(\cdot)$. The flexibility of $g$ makes obtained binocular feature $G_b^k$ take different shapes depending on the extraction operation. We give some examples as follows:
\begin{equation}
G_b^k=
\begin{cases}
g(F_b^k),&\text{in SAT-SE, where}\ G_b^k\in \mathbb{R}^{C\times 1\times 1}\\ 
g(F_b^k),&\text{spatial attention in SAT-CBAM,}\\
&\text{where}\ G_b^k\in \mathbb{R}^{2\times H\times W}\\
g(F_b^k,\omega^g),&\text{in SAT-GC, where}\ G_b^k\in \mathbb{R}^{C\times 1\times 1}
\end{cases}
\end{equation}

\subsubsection{\textbf{Non-Linear Fitting}}
This part generally transforms the gathered binocular features $G_b^k$ into two higher-level attention maps using a series of FC layers and non-linear activations. Instead of an original sigmoid function (SE and CBAM), a softmax function is substituted to generate two sum-to-one weights $W_l^k$ and $W_r^k$ for two monocular branches. As for the attention module (GC) whose last layer was originally designed as FC, we assign the output dimension of FC as $2$ to generate $W_l^k$ and $W_r^k$. The derivation process can be described as
\begin{equation}
    (W_l^k,W_r^k)=w(G_b^k,\omega^w),\ W_l^k+W_r^k=1,
\end{equation}
where $w(\cdot,\omega^w)$ denotes the fitting function and $\omega^w$ is the used parameter.

\subsubsection{\textbf{Top-Down Modulation}}
In this component, previously generated higher-level attention maps $W_l^k$ and $W_r^k$ are utilized as binocular top-down modulation signals to guide the recalibration of two lower-level monocular features $F_l^k$ and $F_r^k$. The procedure of modulation is simple and convenient: 
\begin{equation}
    \tilde{F}_m^k=W_m^k\circledast F_m^k,\ m\in \{l,r\},
\end{equation}
where $\tilde{F}_m^k$ is the recalibrated monocular feature as well as the output of SAT block. ``$\circledast$'' represents the fusion operation. When the design is scaled dot-product attention, ``$\circledast$'' performs element-wise multiplication (SAT-SE and SAT-CBAM) and summation otherwise (SAT-GC).

It is worthwhile noting that during the inference stage, once the input stereo pairs change, $\alpha$ and $W_m^k$ in different phases also change. Experiment results prove that it is especially effective in tackling asymmetric distortions, illustrating its inherent philosophy.

\subsection{Dual-Pooling \& Regression}
At the end of monocular branches and the last SAT block, a convolution layer with $1\times1$ kernel size decodes their outputs. Further, we process the two abstract monocular features.

As done in most NR-SIQA methods~\cite{Zhouwei2019TIP,Liu2020NC,LiuYun-Access,Bourbia2021ICIP,Si2022TIP}, we utilize summation and subtraction operations on two recalibrated high-level monocular features to get fusion and difference maps, which contain much parallax information. We then apply dual-pooling on them to reduce parameters and screen the most crucial quality information for regression. Specifically, the fusion maps are $4\times4$ min-pooled, while the difference maps are $4\times4$ max-pooled. Finally, we flatten the obtained two $5\times5$ abstract quality features and concatenate them together. Three FC layers regress them to a predicted quality score of the stereo pair patches. Moreover, the first two FC layers are each followed by a ReLU function and a dropout unit~\cite{dropout} (dropout rate $=0.5$), aiming to activate neurons and prevent overfitting.

\section{Experiments and Results}
\label{sec:experi}

\subsection{Database}
To evaluate the performance of the proposed method for NR-SIQA, we conducted our experiments on four public 3D databases: LIVE 3D Phase I~\cite{LIVEI}, LIVE 3D Phase II~\cite{LIVEII-FR,LIVEII-NR}, WIVC 3D Phase I~\cite{WIVCI}, and WIVC 3D Phase II\cite{WIVCII}. A detailed introduction of four databases is given below:

\subsubsection{LIVE 3D Phase I}
This database contains five types of distortion: Gaussian Blur (Blur), Fast-Fading (FF), JPEG2000 (JP2K) compression, JPEG compression (JPEG), and Additive White Gaussian Noise (WN). The dataset consists of 20 reference images and 365 distorted images (80 pairs each for JPEG, JP2K, WN, FF, and 45 for Blur) with co-registered human scores in the form of Differential Mean Opinion Score (DMOS). All distortions are symmetric. Depth maps and disparity maps are also provided for reference images.

\subsubsection{LIVE 3D Phase II}
Eight reference images and 360 distorted images with co-registered human scores in the form of DMOS are contained in this database. The distortion types are the same as in LIVE 3D Phase I. For each distortion type, every reference stereo pair was processed to create three symmetric distorted stereo pairs and six asymmetric distorted stereo pairs. In total, there are 120 symmetrically and 240 asymmetrically distorted stereo images. Depth maps and disparity maps of reference images are also given.

\subsubsection{WIVC 3D Phase I}
This database provides six pristine stereo pairs and their corresponding single-view images, which are employed to generate distorted stereo pairs, either symmetrically or asymmetrically. Each single-view image is altered by three types of distortions (WN, Blur, and JPEG), and each distortion type has four distortion levels. Altogether, there are 324 distorted stereo images with their Mean Opinion Score (MOS).

\subsubsection{WIVC 3D Phase II}
The database is created from 10 pristine stereo image pairs. Like the WIVC 3D Phase I database, each single-view image of the reference pair is processed by the same three distortion types and four distortion levels. In total, there are 450 symmetric and asymmetric distorted stereo images with their co-registered MOS values.

\subsection{Performance Index}
In this paper, we employ three indicators, including Pearson Linear Correlation Coefficient (PLCC), Spearman Rank Order Correlation Coefficient (SROCC), and Root Mean Squared Error (RMSE), for performance comparison. PLCC reflects the linear correlation between prediction and ground truth, and SROCC measures the rank correlation between objective and subjective scores. RMSE reflects the prediction accuracy. The floating range of the three indices is [0,1]. When the values of PLCC and SROCC are closer to 1, the RMSE is closer to 0, resulting in a better prediction. Moreover, we perform a non-linear mapping suggested in~\cite{IQA-non-linear} when calculating the metrics. The five-parameter logistic function can be described as:
\begin{equation}
f(Q)=\beta_1(\frac{1}{2}-\frac{1}{1+e^{[\beta_2(Q-\beta_3)]}})+\beta_4Q+\beta_5,
\end{equation}
where $Q$ is the prediction score and $f(Q)$ is the corresponding mapped objective quality score. The $\beta_1,\beta_2,\ldots,\beta_5$ are parameters to be fitted by minimizing the sum of squared errors between the mapped objective score $f(Q)$ and the ground truth.

\subsection{Network Training}
For data pre-processing, we crop raw distorted stereo pairs in four databases to non-overlapping $40\times40$ patches to solve the data deficiency problem. We assign the subjective score of source images as the label of corresponding patch pairs and feed $64$ patches per mini-batch to the model. PyTorch is used for implementation on an NVIDIA GeForce RTX 4090 GPU with 24GB memory. We utilize the Adam~\cite{Adam} optimizer with the default momentum values of $(0.9, 0.999)$ for ($\beta_1$ and $\beta_2$). The weight decay is set to $0.0004$. The initial learning rate is $1\times10^{-4}$, and we adopt cosine decay with warm restarts~\cite{loshchilov2017sgdr} to adjust it for $100$ epochs. We randomly divide $80\%$ of the distorted image pairs as the training set and the rest $20\%$ as the testing set for each database. The k-fold cross-validation ($k=10$ in our experiments) is carried out over four datasets, and the mean value is adopted. The Euclidean distance is applied to the training process. The loss function can be formulated as:
\begin{equation}
\text{Loss} = \frac{1}{N}\sum_{n=1}^{N}\lVert f(x_n;\omega) - y_n\rVert^{2}_2,
\end{equation}
where $N$ is the number of training patch pairs in a batch, $f(x_n;\omega)$ is the predicted quality score of the $n$-th input pairs $x_n$, and $y_n$ is the corresponding ground truth.

\begin{table*}[t]
    \caption{Performance comparison on the two LIVE 3D databases.}
    \label{tab:LIVE}
    \centering
    \begin{tabular}{*{8}{c}}
        \toprule      
        \multirow{2}{*}{}&\multirow{2}{*}{\textbf{Methods}}&\multicolumn{3}{c}{\textbf{LIVE 3D Phase I}}&\multicolumn{3}{c}{\textbf{LIVE 3D Phase II}} \\
        \cmidrule(lr){3-5}\cmidrule(lr){6-8}
        ~ & ~ & \textbf{PLCC} & \textbf{SROCC} & \textbf{RMSE} & \textbf{PLCC} & \textbf{SROCC} & \textbf{RMSE} \\
        \midrule
        \multirow{3}{*}{\textbf{Traditional models}} & Chen~\cite{LIVEII-NR} & 0.895 & 0.891 & 7.247 & 0.880 & 0.880 & 5.102 \\
        ~ & Zhou2016~\cite{ZhouWujie-KNN} & 0.928 & 0.887 & 6.025 & 0.861 & 0.823 & 5.779 \\
        ~ & Yang~\cite{YangJiachen-IoT} & 0.9364 & 0.9289 & 5.2575 & 0.9131 & 0.8747 & 5.0994 \\
        ~ & Messai2020~\cite{Messai-AdaBoost} & 0.939 & 0.930 & 5.605 & 0.922 & 0.913 & 4.352 \\
        ~ & Liu~\cite{Liu2020NC} & 0.958 & 0.949 & 5.069 & 0.935 & 0.933 & 4.014 \\
        \midrule
        \multirow{9}{*}{\textbf{CNN-based models}} & Zhang~\cite{Zhang2016} & 0.947 & 0.943 & 5.336 & - & - & - \\
        ~ & Fang~\cite{Fang2019Siamese} & 0.957 & 0.946 & - & 0.946 & 0.934 & - \\
        ~ & Zhou2019~\cite{Zhouwei2019TIP} & 0.973 & 0.965 & \textbf{3.682} & 0.957 & 0.947 & 3.270 \\
        ~ & Shi~\cite{Shi2020PR} & 0.963 & 0.936 & 4.161 & 0.961 & 0.948 & \textbf{2.675} \\
        ~ & Yan~\cite{Yan2020} & 0.962 & 0.950 & - & 0.946 & 0.938 & - \\
        ~ & Sun~\cite{Sun2020TMM} & 0.951 & 0.959 & 4.573 & 0.938 & 0.918 & 3.809 \\
        ~ & Bourbia~\cite{Bourbia2021ICIP} & 0.957 & 0.942 & - & 0.921 & 0.915 & - \\
        ~ & Zhou2021~\cite{ZhouMingyue} & 0.9768 & 0.9678 & 3.8204 & 0.9561 & 0.9577 & 3.5649 \\
        ~ & Shen~\cite{Shen2021} & 0.972 & 0.962 & - & 0.953 & 0.951 & - \\
        ~ & Sim~\cite{Sim2022} & 0.9697 & 0.9622 & 3.9441 & 0.9619 & 0.955 & 3.0422 \\
        ~ & Si~\cite{Si2022TIP} & \textbf{0.9779} & 0.9656 & \textbf{2.6077} & \textbf{0.9717} & 0.9529 & \textbf{2.2771} \\
        \midrule
        \multirow{3}{*}{\textbf{Ours}} & SAT-SE & \textbf{0.9772} & \textbf{0.9739} & 4.0588 & 0.9635 & \textbf{0.9607} & 3.0613 \\
        ~ & SAT-CBAM & \textbf{0.9773} & \textbf{0.9746} & \textbf{3.7831} & \textbf{0.9638} & \textbf{0.9641} & 2.9126 \\
        ~ & SAT-GC & \textbf{0.9772} & \textbf{0.9723} & 3.8718 & \textbf{0.9637} & \textbf{0.9616} & \textbf{2.8951} \\
        \bottomrule      
    \end{tabular}
\end{table*}

\begin{table}[t]
    \caption{Performance comparison on the two WIVC 3D databases.}
    \label{tab:WIVC}
    \centering
    \resizebox{\linewidth}{!}{
    \begin{tabular}{*{7}{c}}
        \toprule      
        \multirow{2}{*}{\textbf{Methods}}&\multicolumn{3}{c}{\textbf{WIVC 3D Phase I}} & \multicolumn{3}{c}{\textbf{WIVC 3D Phase II}} \\
        \cmidrule(lr){2-4}\cmidrule(lr){5-7}
        ~ & \textbf{PLCC} & \textbf{SROCC} & \textbf{RMSE} & \textbf{PLCC} & \textbf{SROCC} & \textbf{RMSE} \\
        \midrule
        Fang~\cite{Fang-Access} & 0.953 & 0.950 & - & 0.936 & 0.922 & - \\
        Yang~\cite{YangJiachen-auto} & 0.9439 & 0.9246 & - & 0.9331 & 0.9143 & - \\
        Liu~\cite{Liu2020NC} & 0.945 & 0.928 & 5.268 & 0.913 & 0.901 & 7.658 \\
        \midrule
        Sun~\cite{Sun2020TMM} & - & - & - & 0.840 & 0.835 & - \\
        Messai2022~\cite{Messai2022ICIP} & 0.972 & 0.967 & 3.635 & 0.971 & 0.966 & 4.161 \\
        Sim~\cite{Sim2022} & 0.9625 & 0.9566 & 4.1916 & 0.9698 & 0.9699 & 4.5981 \\
        Si~\cite{Si2022TIP} & 0.9690 & 0.9599 & \textbf{2.9508} & 0.9580 & 0.9501 & \textbf{3.0506} \\
        \midrule
        SAT-SE & \textbf{0.9725} & \textbf{0.9698} & 3.3265 & \textbf{0.9717} & \textbf{0.9705} & 3.5363 \\
        SAT-CBAM & \textbf{0.9738} & \textbf{0.9740} & \textbf{3.3106} & \textbf{0.9735} & \textbf{0.9734} & \textbf{3.5136} \\
        SAT-GC & \textbf{0.9734} & \textbf{0.9716} & \textbf{3.3051} & \textbf{0.9732} & \textbf{0.9718} & \textbf{3.5260} \\
        \bottomrule      
    \end{tabular}
    }
\end{table}

\subsection{Performance comparison}
\label{subsec:perf-comparison}
We reconstruct three well-known attention blocks with SAT structure for NR-SIQA, including SE~\cite{SE}, CBAM~\cite{CBAM}, and GC~\cite{GC}. For overall performance comparison, we stack $7$ SAT blocks $(K=7)$ for these three variants. We compare our models with sixteen state-of-the-art NR-SIQA methods on LIVE 3D databases, including five traditional~\cite{LIVEII-NR,ZhouWujie-KNN,YangJiachen-IoT,Messai-AdaBoost,Liu2020NC} and eleven CNN-based methods~\cite{Zhang2016,Fang2019Siamese,Zhouwei2019TIP,Shi2020PR,Yan2020,Sun2020TMM,Bourbia2021ICIP,ZhouMingyue,Shen2021,Sim2022,Si2022TIP}. Seven state-of-the-art NR-SIQA methods are compared with our networks on WIVC 3D databases, including three traditional~\cite{Fang-Access,YangJiachen-auto,Liu2020NC} and four CNN-based methods~\cite{Sun2020TMM,Messai2022ICIP,Sim2022,Si2022TIP}. Table~\ref{tab:LIVE} and~\ref{tab:WIVC} demonstrate the comparison results on LIVE 3D databases and WIVC 3D databases, respectively. The results of all comparison methods are directly borrowed from corresponding papers. The top three performances are marked as \textbf{bold}, and unavailable indicators are marked as ``-''.

As shown in Table~\ref{tab:LIVE} and Table~\ref{tab:WIVC}, three SATNet variants outperform conventional NR-SIQA methods and obtain competitive indicators against recent-year CNN-based works. SAT-CBAM obtains the best overall performance on four databases due to its rational attention considerations on channel and spatial dimensions. Generally, most traditional works are inferior to CNN-based ones, which can be attributed to the limitations of hand-crafted features. As a pioneer work of applying CNN to NR-SIQA, Zhang~\cite{Zhang2016} made a great breakthrough compared with other traditional approaches in 2016. Zhou2019~\cite{Zhouwei2019TIP} is a classic dual-stream fashion method, and its design performs better than some research today. Even though Zhou2019~\cite{Zhouwei2019TIP} gained better RMSE than our models, SATNets have more powerful performance in terms of three indicators on the LIVE 3D Phase II database, showcasing that simple multi-level information interaction mode has a problem in tackling asymmetric distortion. Especially, SATNets lead to a considerable improvement in the LIVE 3D Phase II database, WIVC 3D Phase I, and WIVC 3D Phase II database, indicating the validity of top-down modulation from higher-level binocular features to lower-level monocular ones in processing asymmetric distortion.

Table~\ref{tab:LIVE} and~\ref{tab:WIVC} also reveal that StereoIF-Net, recently proposed by Si et al.~\cite{Si2022TIP}, gets the highest PLCC and lowest RMSE in two LIVE 3D databases, which can be credited to its complex structure in pursuit of simulating both binocular interaction and fusion mechanisms of the HVS. At the same time, substantial kernel numbers of convolutional layers are introduced to the intricate binocular interaction module (BIM). High complexity and computational overhead pay for high performance, which will be discussed in more detail in Section~\ref{subsec:complex}. In contrast, our models obtain the highest PLCC and SROCC values on two WIVC 3D databases with lightweight model design and computational cost.

We also perform individual distortion type tests on two LIVE 3D databases to evaluate the competence of our method when handling specific distortion types. Table~\ref{tab:PLCC-LIVE} and Table~\ref{tab:SROCC-LIVE} provide the comparison results between some methods and SAT-SE on two LIVE 3D databases for PLCC and SROCC metrics respectively, and the top two performances are \textbf{bold}. From Table~\ref{tab:PLCC-LIVE} and Table~\ref{tab:SROCC-LIVE}, it is evident that SAT-SE performs remarkably in predicting quality for specific types of distortions, particularly in the cases of WN and BLUR, outperforming other methods significantly. This superiority can be attributed to its sensitivity towards distortion types characterized by such distributions. 

\begin{table*}[t]
    \caption{PLCC comparison of individual distortion types on two LIVE 3D databases.}
    \label{tab:PLCC-LIVE}
    \centering
    \begin{tabular}{*{11}{c}}
        \toprule      
        \multirow{2}{*}{\textbf{Methods}}&\multicolumn{5}{c}{\textbf{LIVE 3D Phase I}}&\multicolumn{5}{c}{\textbf{LIVE 3D Phase II}} \\
        \cmidrule(lr){2-6}\cmidrule(lr){7-11}
        ~ & \textbf{BLUR} & \textbf{FF} & \textbf{JP2K} & \textbf{JPEG} & \textbf{WN} & \textbf{BLUR} & \textbf{FF} & \textbf{JP2K} & \textbf{JPEG} & \textbf{WN} \\
        \midrule
        Chen~\cite{LIVEII-NR} & 0.917 & 0.735 & 0.907 & 0.695 & 0.917 & 0.941 & 0.932 & 0.899 & 0.901 & 0.947 \\
        Messai2020~\cite{Messai-AdaBoost} & 0.935 & 0.845 & 0.926 & 0.668 & 0.941 & 0.978 & 0.925 & 0.835 & 0.859 & 0.953 \\
        Liu~\cite{Liu2020NC} & 0.956 & 0.855 & 0.938 & 0.810 & 0.966 & 0.987 & 0.959 & 0.936 & 0.867 & 0.969 \\
        Zhang~\cite{Zhang2016} & 0.930 & 0.883 & 0.926 & 0.740 & 0.944 & - & - & - & - & - \\
        Fang~\cite{Fang2019Siamese} & 0.953 & 0.868 & 0.975 & 0.753 & 0.973 & 0.983 & 0.929 & \textbf{0.975} & \textbf{0.952} & \textbf{0.972} \\    
        Zhou2019~\cite{Zhouwei2019TIP} & 0.974 & \textbf{0.965} & \textbf{0.988} & \textbf{0.916} & \textbf{0.988} & 0.955 & \textbf{0.994} & 0.905 & \textbf{0.933} & \textbf{0.972} \\
        Sun~\cite{Sun2020TMM} & 0.960 & 0.890 & 0.948 & 0.806 & 0.956 & \textbf{0.996} & 0.901 & 0.900 & 0.823 & 0.956 \\
        Bourbia~\cite{Bourbia2021ICIP} & 0.972 & 0.838 & 0.951 & 0.744 & 0.966 & 0.986 & 0.948 & 0.912 & 0.874 & 0.924 \\
        Zhou2021~\cite{ZhouMingyue} & \textbf{0.9885} & 0.9315 & 0.9601 & 0.9154 & 0.9796 & 0.9745 & 0.9601 & 0.9221 & 0.8968 & 0.9432 \\
        Shen~\cite{Shen2021} & 0.988 & 0.939 & \textbf{0.984} & 0.906 & 0.947 & 0.988 & \textbf{0.964} & \textbf{0.956} & 0.825 & 0.954 \\
        SAT-SE & \textbf{0.9956} & \textbf{0.9583} & 0.9645 & \textbf{0.9346} & \textbf{0.9959} & \textbf{0.9970} & 0.9605 & 0.9363 & 0.9066 & \textbf{0.9992} \\
        \bottomrule      
    \end{tabular}
\end{table*}

\begin{table*}[t]
    \caption{SROCC comparison of individual distortion types on two LIVE 3D databases.}
    \label{tab:SROCC-LIVE}
    \centering
    \begin{tabular}{*{11}{c}}
        \toprule      
        \multirow{2}{*}{\textbf{Methods}}&\multicolumn{5}{c}{\textbf{LIVE 3D Phase I}}&\multicolumn{5}{c}{\textbf{LIVE 3D Phase II}} \\
        \cmidrule(lr){2-6}\cmidrule(lr){7-11}
        ~ & \textbf{BLUR} & \textbf{FF} & \textbf{JP2K} & \textbf{JPEG} & \textbf{WN} & \textbf{BLUR} & \textbf{FF} & \textbf{JP2K} & \textbf{JPEG} & \textbf{WN} \\
        \midrule
        Chen~\cite{LIVEII-NR} & 0.878 & 0.652 & 0.863 & 0.617 & 0.919 & 0.900 & 0.933 & 0.867 & 0.867 & 0.950 \\
        Messai2020~\cite{Messai-AdaBoost} & 0.887 & 0.777 & 0.899 & 0.625 & 0.941 & 0.913 & 0.925 & 0.842 & 0.837 & 0.943 \\
        Liu~\cite{Liu2020NC} & 0.917 & 0.821 & 0.888 & 0.785 & 0.951 & 0.936 & 0.938 & 0.909 & 0.825 & 0.946 \\
        Zhang~\cite{Zhang2016} & 0.909 & 0.834 & 0.931 & 0.693 & 0.946 & - & - & - & - & - \\
        Fang~\cite{Fang2019Siamese} & - & - & - & - & - & - & - & - & - & - \\
        Zhou2019~\cite{Zhouwei2019TIP} & 0.855 & \textbf{0.917} & 0.961 & 0.912 & 0.965 & 0.600 & 0.951 & 0.874 & 0.747 & 0.942 \\
        Sun~\cite{Sun2020TMM} & \textbf{0.979} & 0.853 & \textbf{0.970} & 0.687 & 0.893 & \textbf{0.964} & 0.918 & 0.897 & 0.579 & 0.933 \\
        Bourbia~\cite{Bourbia2021ICIP} & 0.867 & 0.782 & 0.908 & 0.679 & 0.949 & 0.924 & 0.933 & 0.915 & \textbf{0.874} & 0.889 \\
        Zhou2021~\cite{ZhouMingyue} & \textbf{0.9790} & \textbf{0.9412} & 0.9429 & \textbf{0.9385} & \textbf{0.9824} & 0.9333 & \textbf{0.9650} & 0.9414 & \textbf{0.8681} & \textbf{0.9587} \\
        Shen~\cite{Shen2021} & 0.945 & 0.900 & \textbf{0.965} & 0.879 & 0.921 & 0.951 & \textbf{0.969} & \textbf{0.954} & 0.816 & 0.923 \\
        SAT-SE & \textbf{0.9944} & 0.9097 & 0.9514 & \textbf{0.9243} & \textbf{0.9965} & \textbf{0.9797} & 0.9585 & \textbf{0.9423} & \textbf{0.8740} & \textbf{0.9839} \\
        \bottomrule      
    \end{tabular}
\end{table*}

\begin{table}[t]
    \caption{Performance comparison of cross-dataset validation on two LIVE 3D databases.}
    \label{tab:cross-LIVE}
    \centering
    \begin{tabular}{*{5}{c}}
        \toprule      
        \multirow{2}{*}{\textbf{Methods}}&\multicolumn{2}{c}{\textbf{LIVE 3D Phase I/II}} & \multicolumn{2}{c}{\textbf{LIVE 3D Phase II/I}} \\
        \cmidrule(lr){2-3}\cmidrule(lr){4-5}
        ~ & \textbf{PLCC} & \textbf{SROCC} & \textbf{PLCC} & \textbf{SROCC} \\
        \midrule
        Messai2020~\cite{Messai-AdaBoost} & 0.832 & 0.823 & 0.897 & 0.887 \\
        Liu~\cite{Liu2020NC} & \textbf{0.862} & 0.832 & 0.888 & 0.874 \\
        Fang~\cite{Fang2019Siamese} & 0.811 & 0.797 & 0.899 & 0.898 \\
        Zhou2019~\cite{Zhouwei2019TIP} & 0.710 & - & \textbf{0.932} & - \\
        Shi~\cite{Shi2020PR} & - & 0.793 & - & 0.901 \\
        Shen~\cite{Shen2021} & 0.848 & \textbf{0.833} & 0.915 & \textbf{0.921} \\
        Sim~\cite{Sim2022} & 0.8041 & 0.7704 & 0.9083 & 0.8964 \\
        Si~\cite{Si2022TIP} & \textbf{0.8595} & 0.7972 & 0.9184 & 0.9149 \\
        SAT-SE & 0.8547 & \textbf{0.8378} & \textbf{0.9204} & \textbf{0.9187} \\
        \bottomrule      
    \end{tabular}
\end{table}

\begin{table}[t]
    \caption{Performance comparison of cross-dataset validation on two WIVC 3D databases.}
    \label{tab:cross-WIVC}
    \centering
    \begin{tabular}{*{5}{c}}
        \toprule      
        \multirow{2}{*}{\textbf{Methods}}&\multicolumn{2}{c}{\textbf{WIVC 3D Phase I/II}} & \multicolumn{2}{c}{\textbf{WIVC 3D Phase II/I}} \\
        \cmidrule(lr){2-3}\cmidrule(lr){4-5}
        ~ & \textbf{PLCC} & \textbf{SROCC} & \textbf{PLCC} & \textbf{SROCC} \\
        \midrule
        Yang~\cite{YangJiachen-auto} & \textbf{0.8421} & \textbf{0.8313} & 0.8739 & 0.8687 \\
        Liu~\cite{Liu2020NC} & 0.696 & 0.627 & 0.701 & 0.708 \\
        Sim~\cite{Sim2022} & 0.7876 & 0.7547 & \textbf{0.9139} & \textbf{0.9032} \\
        Si~\cite{Si2022TIP} & 0.7816 & 0.7502 & 0.8981 & 0.8712 \\
        SAT-SE & \textbf{0.8065} & \textbf{0.7808} & \textbf{0.9216} & \textbf{0.9126} \\
        \bottomrule      
    \end{tabular}
\end{table}

\subsection{Cross-dataset validation}
In order to verify the generalization and robustness of the proposed method, cross-database validation experiments on two LIVE 3D databases and two WIVC 3D databases are carried out respectively. The ``Training set/Test set'' experimental results are illustrated in Table~\ref{tab:cross-LIVE} and Table~\ref{tab:cross-WIVC}, the top two metric values are marked as \textbf{bold} and unavailable indicators are marked as ``-''. For example, ``LIVE 3D Phase I/II'' indicates that the LIVE 3D Phase I database is taken as the training set, and the LIVE 3D Phase II database is used as the test set.

\begin{table*}[t]
    \caption{Performance comparison of different model scales on two LIVE 3D databases.}
    \label{tab:abla-scale}
    \centering
    \begin{tabular}{*{10}{c}}
        \toprule      
        \multirow{2}{*}{\makecell[c]{\textbf{Block}\\\textbf{number ($K$)}}}&\multirow{2}{*}{\makecell{\textbf{Which Res Block}\\\textbf{to employ}}}&\multirow{2}{*}{\makecell{\textbf{Depth}\\\textbf{(layers)}}}&\multirow{2}{*}{\textbf{\#Param.}}&\multicolumn{3}{c}{\textbf{LIVE 3D Phase I}}&\multicolumn{3}{c}{\textbf{LIVE 3D Phase II}} \\
        \cmidrule(lr){5-7}\cmidrule(lr){8-10}
        ~ & ~ & ~ & ~ & \textbf{PLCC} & \textbf{SROCC} & \textbf{RMSE} & \textbf{PLCC} & \textbf{SROCC} & \textbf{RMSE} \\
        \midrule
        $K=3$ & Basic Block & 11 & 6.87M & 0.9758 & \textbf{0.9751} & \textbf{3.8932} & 0.9615 & 0.9590 & \textbf{2.9347} \\
        $K=7$ (Proposed) & Basic Block & 19 & 7.47M & \textbf{0.9772} & 0.9739 & 4.0588 & 0.9635 & 0.9607 & 3.0613 \\
        $K=14$ & Basic Block & 33 & 8.51M & 0.9741 & 0.9703 & 4.5037 & \textbf{0.9644} & \textbf{0.9628} & 2.9466 \\
        $K=15$ & Bottleneck & 50 & 7.80M & 0.9743 & 0.9733 & 4.6435 & 0.9640 & 0.9593 & 2.9388 \\
        \bottomrule      
    \end{tabular}
\end{table*}

It can be found that the experimental results of cross-dataset validation are not as good as those in the case that the data of the training and test set come from the same database. Yet, our model still has competitive prediction performance over LIVE 3D databases and WIVC 3D databases against other methods. Although PLCC of Liu~\cite{Liu2020NC} and Si~\cite{Si2022TIP} on LIVE 3D Phase I/II are slightly better than ours, SROCC of SAT-SE outperforms both methods, as well as on other ``Training/Test'' patterns. Yang~\cite{YangJiachen-auto} reap the best performance on WIVC 3D Phase I/II, but on WIVC 3D Phase II/I, their work significantly underperformed ours and SAT-SE achieves the top one. It is worth highlighting that in both Table~\ref{tab:cross-LIVE} and Table~\ref{tab:cross-WIVC}, SAT-SE occupies the highest number of bold values, which reflects the excellent generality of the proposed algorithm.

For both LIVE 3D databases and WIVC 3D databases, the ``Phase II/I'' pattern consistently yields superior results to the ``Phase I/II'' pattern, as the complexity of distorted images in Phase II surpasses that of Phase I. In addition, the LIVE 3D Phase II dataset contains stereo pairs distorted symmetrically and asymmetrically, while LIVE 3D Phase I contains only stereo pairs with symmetric distortions. WIVC 3D Phase II database involves more scenes and distorted images than WIVC 3D Phase I database, i.e., more training data have led to better results with the same data complexity.

\subsection{Ablation Study}
The experiments of the ablation study are conducted based on SAT-SE implementation in view of its efficiency. The best performance of each experiment is highlighted in \textbf{boldface}.

\subsubsection{\textbf{Choice of the model scale of SATNet for NR-SIQA}}
As mentioned in Section~\ref{subsec:MonoBran}, we construct our model with different scales with a range of SAT block numbers ($K$). Here, we assign $K=3,7,14,15$, corresponding to four scales models (depth $=11,19,33,50$, respectively) to investigate the most cost-effective $K$ for the NR-SIQA task. The configurations and performances of different scales of SATNet are shown in Table~\ref{tab:abla-scale}.

A bottleneck layer can significantly reduce the computational cost when increasing the network depth, which is exactly the reason why the 50-layer model has fewer parameters than the 33-layer model (7.80M versus 8.51M). From experimental results, we observe that with more SAT blocks stacked, SATNet goes deeper and performs better in general. Inevitably, the deep model incurs more redundant parameters, manifesting overfitting and performance degradation on both databases. Notably, even shallower models ($K=3,7$) attain the best PLCC, SROCC, and RMSE on LIVE 3D Phase I. Consequently, we choose $K=7$ for overall experiments in pursuit of the balance between performance improvement and computational overhead.

\begin{table}[t]
    \caption{Performance comparison of design perspectives on two LIVE 3D databases.}
    \label{tab:Top-Down}
    \centering
    \begin{tabular}{*{5}{c}}
        \toprule      
        \multirow{2}{*}{\textbf{Methods}}&\multicolumn{2}{c}{\textbf{LIVE 3D Phase I}} & \multicolumn{2}{c}{\textbf{LIVE 3D Phase II}} \\
        \cmidrule(lr){2-3}\cmidrule(lr){4-5}
        ~ & \textbf{PLCC} & \textbf{SROCC} & \textbf{PLCC} & \textbf{SROCC} \\
        \midrule
        bottom-up-1 & 0.9563 & 0.9401 & 0.9445 & 0.9321 \\
        bottom-up-2 & 0.9584 & 0.9440 & 0.9564 & 0.9423 \\
        bottom-up-3 & 0.9625 & 0.9544 & 0.9529 & 0.9420 \\
        top-down (SAT-SE) & \textbf{0.9772} & \textbf{0.9739} & \textbf{0.9635} & \textbf{0.9607} \\
        \bottomrule      
    \end{tabular}
\end{table}

\begin{table}[t]
    \caption{Performance comparison of EC on two LIVE 3D databases.}
    \label{tab:abla-EC}
    \centering
    \begin{tabular}{*{5}{c}}
        \toprule      
        \multirow{2}{*}{\textbf{Methods}}&\multicolumn{2}{c}{\textbf{LIVE 3D Phase I}} & \multicolumn{2}{c}{\textbf{LIVE 3D Phase II}} \\
        \cmidrule(lr){2-3}\cmidrule(lr){4-5}
        ~ & \textbf{PLCC} & \textbf{SROCC} & \textbf{PLCC} & \textbf{SROCC} \\
        \midrule
        without EC & 0.9592 & 0.9531 & 0.9489 & 0.9465 \\
        SAT-SE & \textbf{0.9772} & \textbf{0.9739} & \textbf{0.9635} & \textbf{0.9607} \\
        \bottomrule      
    \end{tabular}
\end{table}

\begin{table}[t]
    \caption{Performance comparison of different pooling strategies on the LIVE 3D Phase II database.}
    \label{tab:abla-pool}
    \centering
    \begin{tabular}{*{5}{c}}
        \toprule      
        \multicolumn{3}{c}{\textbf{Pooling strategies}} & \multirow{2}{*}{\textbf{PLCC}} & \multirow{2}{*}{\textbf{SROCC}} \\
        \cmidrule(lr){1-3}
        \textbf{Avg} & \textbf{Max} & \textbf{Min} & ~ & ~ \\
        \midrule
        $+-$ & ~ & ~ & 0.9465 & 0.9421 \\
        ~ & $+-$ & ~ & 0.9588 & 0.9579 \\
        ~ & ~ & $+-$ & 0.9503 & 0.9487 \\
        ~ & $+$ & $-$ & 0.9576 & 0.9543 \\
        ~ & $-$ & $+$ & \textbf{0.9635} & \textbf{0.9607} \\
        \bottomrule      
    \end{tabular}
\end{table}

\begin{table}[t]
    \caption{Computational complexity comparison on the LIVE 3D Phase I database.}
    \label{tab:complex}
    \centering
    \begin{tabular}{*{3}{c}}
        \toprule      
        \textbf{Methods} & \textbf{\#Param.} & \textbf{Inference time (sec)} \\
        \midrule
        Si~\cite{Si2022TIP} & 3.83G & - \\
        SAT-SE & \textbf{7.47M} & \textbf{0.0060} \\
        SAT-CBAM & \textbf{7.47M} & 0.0073 \\
        SAT-GC & \textbf{7.47M} & 0.0063 \\
        \bottomrule      
    \end{tabular}
\end{table}

\subsubsection{\textbf{Verification of top-down philosophy}}
In order to verify the inherent superiority of the top-down perspective, a series of comparison experiments of three different model designs are also conducted. The results are reported in Table~\ref{tab:Top-Down}. As illustrated in Fig.~\ref{fig:Top-Down}, there are two classic model types for NR-SIQA from a bottom-up design perspective: a network without any (explicitly) interaction between monocular information or between binocular and monocular information (or both) during feature extraction; a multi-level network where the fusion or difference features (or both) are obtained as the interactive information. Accordingly, we remove all SAT blocks to implement the first type on SATNet and denote it by ``bottom-up-1''. ``bottom-up-2'' indicates that fusion and difference maps are obtained after the first, third, fifth, and seventh residual blocks and concatenated before regression based on ``bottom-up-1''. Moreover, we substitute SAT blocks in SATNet with vanilla attention blocks to conduct a comparison experiment labeled ``bottom-up-3''. Specifically, two outputs of monocular branches are concatenated and fed into the vanilla attention module. Then, re-weighted higher-level binocular features are divided into two parts again in the channel dimension and forwarded to the original monocular branches instead of being used to modulate lower-level monocular features, which remain bottom-up perspective.

It can be found in Table~\ref{tab:Top-Down} that ``bottom-up-2'' and ``bottom-up-3'' both improve the performance against ``bottom-up-1'', signifying the vital influence of multi-level interaction of information. The proposed SATNet achieves the best performance among all competitors, which can prove the effectiveness of the top-down design philosophy.

\subsubsection{\textbf{Effects of Energy Coefficient}}
\begin{figure*}[t]
\centering
\resizebox{\linewidth}{!}{
\subfloat[On LIVE 3D Phase I.]{\label{subfig:EC-LIVEI}
\begin{minipage}[t]{0.25\linewidth}
\includegraphics[width=\linewidth]{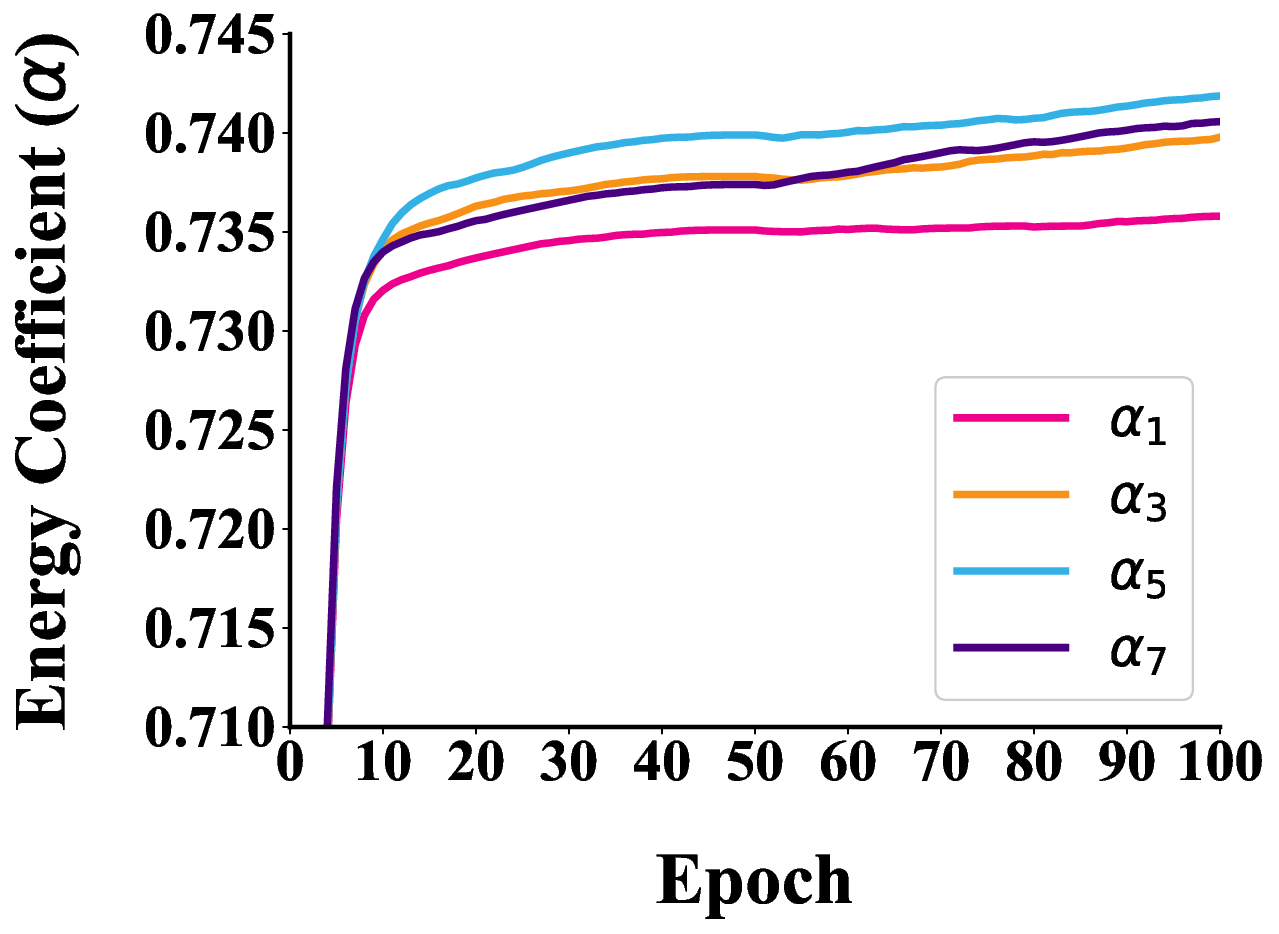}
\end{minipage}
}
\subfloat[On LIVE 3D Phase II.]{\label{subfig:EC-LIVEII}
\begin{minipage}[t]{0.25\linewidth}
\includegraphics[width=\linewidth]{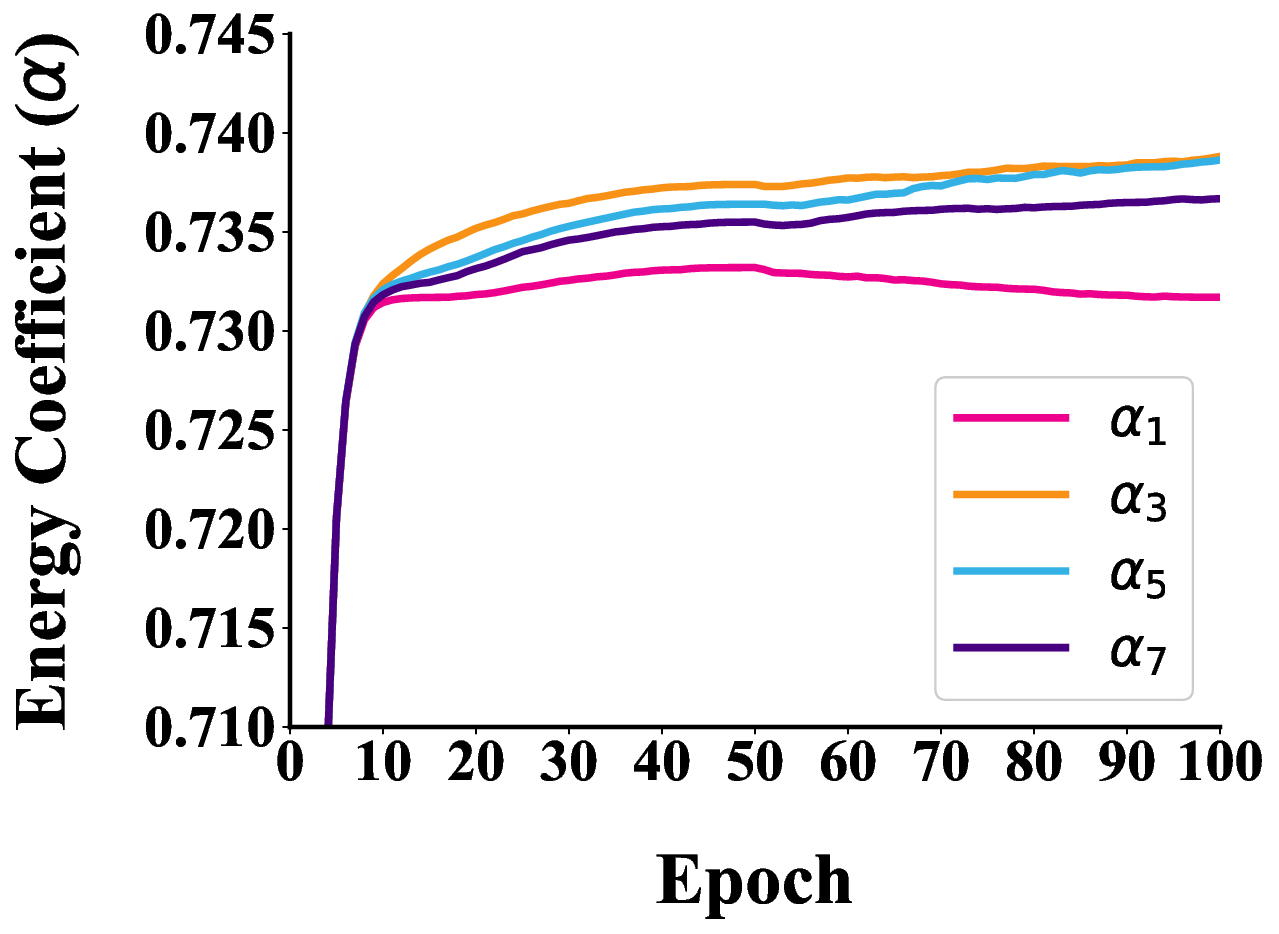}
\end{minipage}
}
\subfloat[On WIVC 3D Phase I.]{\label{subfig:EC-WIVCI}
\begin{minipage}[t]{0.25\linewidth}
\includegraphics[width=\linewidth]{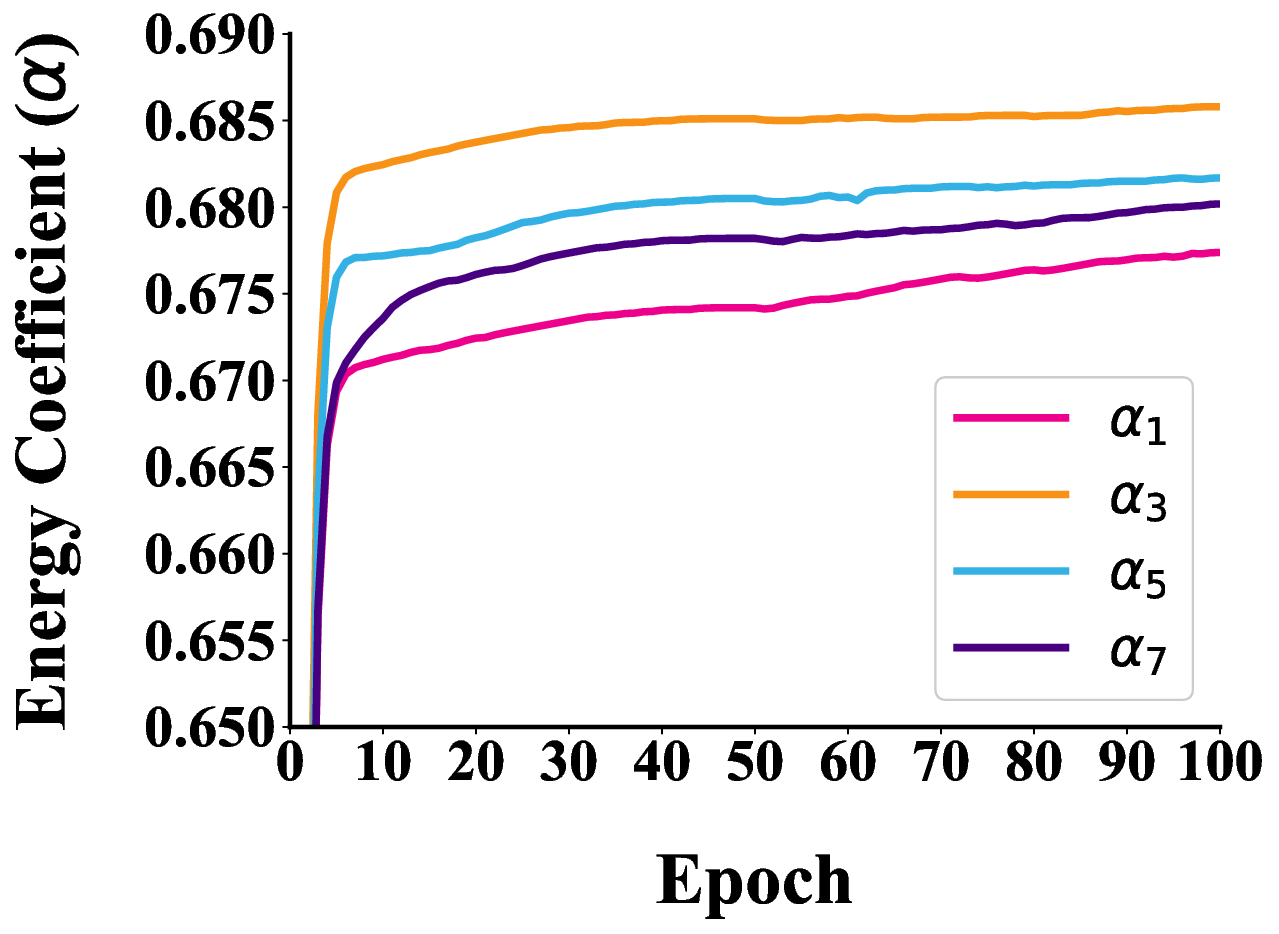}
\end{minipage}
}
\subfloat[On WIVC 3D Phase II.]{\label{subfig:EC-WIVCII}
\begin{minipage}[t]{0.25\linewidth}
\includegraphics[width=\linewidth]{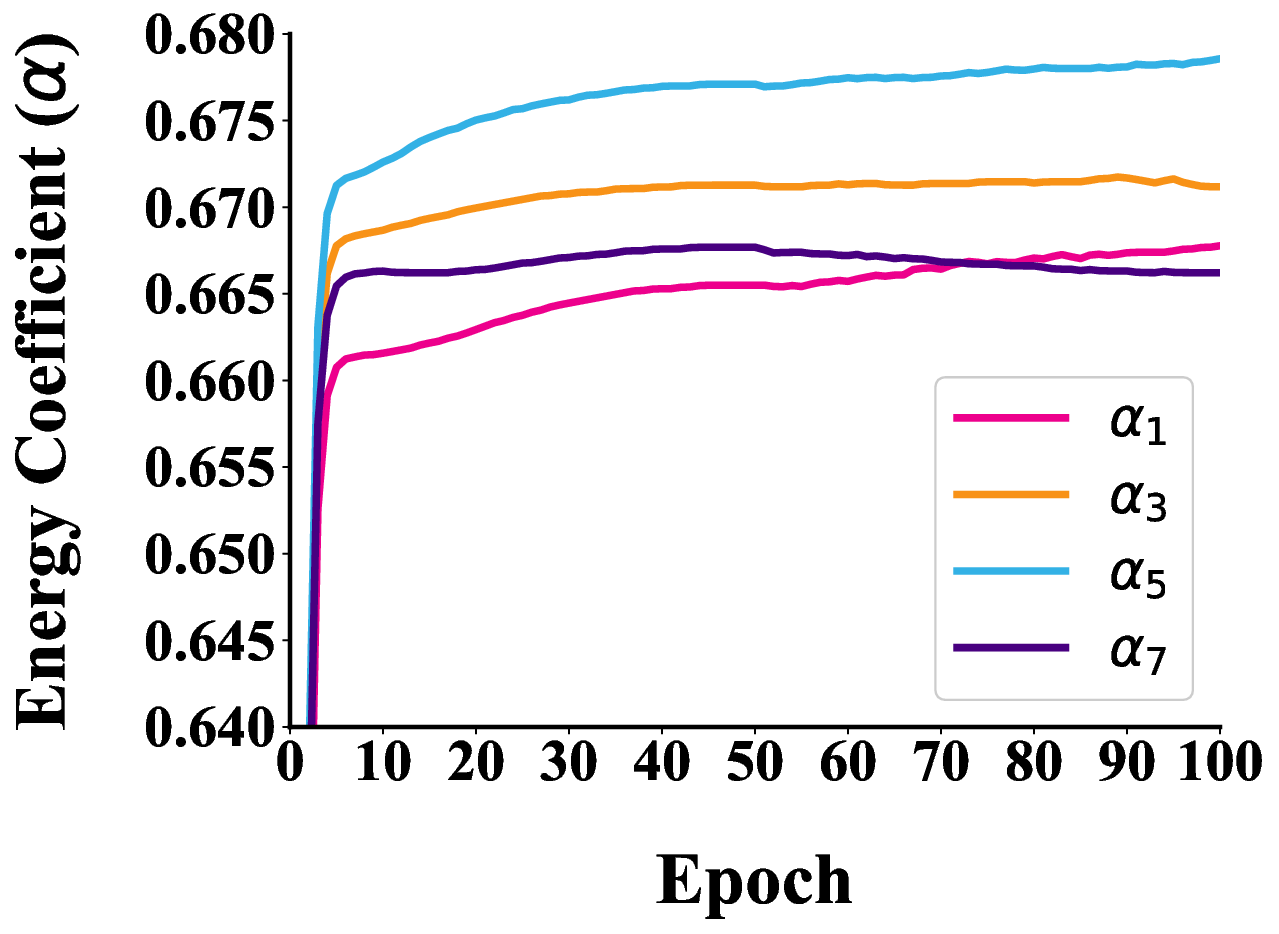}
\end{minipage}
}
}
\caption{Learning processes of Energy Coefficients ($\alpha$) in SAT-SE of different phases on four 3D databases.}
\label{fig:EC}
\end{figure*}

We visualize four ECs' updating processes during training on four databases in Fig.~\ref{fig:EC}. $\alpha_k$ represents the corresponding EC in the $k$ th SAT-SE block. We can observe that the four curves converge quickly and maintain stability in their values individually after about ten epochs. The same phase's EC varies among different databases, and ECs of different phases on the same database are also diverse, which reflects that EC can adjust the magnitude of binocular response adaptively. We also conduct comparison experiments that ablate EC from SAT-SE on two LIVE 3D databases and reveal results in Table~\ref{tab:abla-EC}. Experimental results show that the introduced EC improves PLCC and SROCC significantly on two LIVE 3D databases compared without multiplying EC, which further testifies to the improvement of the proposed EC.

\subsubsection{\textbf{Verification of dual-pooling strategy}}
To verify the superiority of the proposed dual-pooling strategy, we conduct several comparison experiments by adopting some common pooling manners, e.g., average pooling (AVG), max pooling (Max), and min pooling (Min). Results are revealed in Table~\ref{tab:abla-pool}. ``$+$'' and ``$-$'' in the table remark fusion and difference maps, respectively. For instance, the first row of Table~\ref{tab:abla-pool} denotes that both fusion and difference features are processed by average pooling operation before regression since both ``$+$'' and ``$-$'' symbols are in the ``Avg'' column. We can find that the use of average pooling leads to performance deterioration. Although average pooling excels in integrating contextual information, it will smooth learned sharp quality features and cannot get desirable results. Most works generally operate max pooling on both fusion and difference maps (the second row of Table~\ref{tab:abla-pool}), while it can be seen from the indices that our dual-pooling strategy (the last row of Table~\ref{tab:abla-pool}) can filter crucial distortion information for quality regression and achieve the best performance.

\subsection{Computational Complexity}
\label{subsec:complex}
In this subsection, we test the parameters and inference speed of our three SATNet variants to demonstrate the computational complexity. As Section~\ref{subsec:perf-comparison} mentioned, StereoIF-Net~\cite{Si2022TIP} gains top performance on some indices due to its complicated structure and high computational complexity, so we also choose this model as a comparison. Given that the source code of StereoIF-Net is unavailable, we reproduce it and implement the inference based on the same server (NVIDIA GeForce RTX 4090 GPU with 24GB memory) and settings as ours for a fair comparison. Table~\ref{tab:complex} shows the parameters of the models and their average inference time (sec) of one stereo image on the LIVE 3D Phase I database. 

It can be found that the parameters of the three SATNet variants are nearly identical (7.47M) since three FC layers with 6.40M neurons dominate the model's scale instead of light-weight SAT blocks. It only takes 0.0060 seconds for the most efficient SAT-SE to test one stereo image on the LIVE 3D Phase I database, i.e., our SATNet can evaluate 166 stereo images within one second, which can satisfy the requirement of real-time processing of 3D movies. For comparison, large channel numbers in BIM bring enormous parameters (3.83G) to StereoIF-Net, which are approximately 512 times ours. We are unable to train and test this huge model on our server due to the limited 24GB GPU memory capacity, so we can not provide the inference time of StereoIF-Net and denote it by ``-''. Comprehensively, our SATNet is much superior to Si's~\cite{Si2022TIP} in pursuing the balance between model performance and efficiency.

\section{Conclusion}
\label{sec:conclu}
In conclusion, our work contributes to the field of SIQA, which introduces a top-down perspective to guide the assessment by customizing components and input/output of SAT block and incorporating higher-level binocular signals to modulate lower-level monocular features. The proposed adaptive EC enhances the formation of robust binocular features, effectively simulating the process of visual perception. The integration of the dual-pooling strategy further refines the filtering process of quality information, leading to performance improvement. Experimental results prove the philosophy and superiority of our SATNet compared to state-of-the-art methods. Furthermore, we will explore more possibilities for implementing a top-down perspective in 3D quality evaluation and other visual technologies.

\IEEEpubidadjcol



%

\bibliographystyle{IEEEtran}
\bibliography{reference} 

\begin{thebibliography}{10}
\providecommand{\url}[1]{#1}
\csname url@samestyle\endcsname
\providecommand{\newblock}{\relax}
\providecommand{\bibinfo}[2]{#2}
\providecommand{\BIBentrySTDinterwordspacing}{\spaceskip=0pt\relax}
\providecommand{\BIBentryALTinterwordstretchfactor}{4}
\providecommand{\BIBentryALTinterwordspacing}{\spaceskip=\fontdimen2\font plus
\BIBentryALTinterwordstretchfactor\fontdimen3\font minus
  \fontdimen4\font\relax}
\providecommand{\BIBforeignlanguage}[2]{{%
\expandafter\ifx\csname l@#1\endcsname\relax
\typeout{** WARNING: IEEEtran.bst: No hyphenation pattern has been}%
\typeout{** loaded for the language `#1'. Using the pattern for}%
\typeout{** the default language instead.}%
\else
\language=\csname l@#1\endcsname
\fi
#2}}
\providecommand{\BIBdecl}{\relax}
\BIBdecl

\bibitem{ZhouWujie-KNN}
W.~Zhou and L.~Yu, ``Binocular responses for no-reference 3d image quality
  assessment,'' \emph{IEEE Transactions on Multimedia}, vol.~18, no.~6, pp.
  1077--1084, 2016.

\bibitem{ZhouWujie-ELM}
W.~Zhou, L.~Yu, Y.~Zhou, W.~Qiu, M.-W. Wu, and T.~Luo, ``Blind quality
  estimator for 3d images based on binocular combination and extreme learning
  machine,'' \emph{Pattern Recognition}, vol.~71, pp. 207--217, 2017.

\bibitem{Fang-Access}
Y.~Fang, J.~Yan, J.~Wang, X.~Liu, G.~Zhai, and P.~Le~Callet, ``Learning a
  no-reference quality predictor of stereoscopic images by visual binocular
  properties,'' \emph{IEEE Access}, vol.~7, pp. 132\,649--132\,661, 2019.

\bibitem{LiuYun-Access}
Y.~Liu, W.~Yan, Z.~Zheng, B.~Huang, and H.~Yu, ``Blind stereoscopic image
  quality assessment accounting for human monocular visual properties and
  binocular interactions,'' \emph{IEEE Access}, vol.~8, pp. 33\,666--33\,678,
  2020.

\bibitem{Messai-AdaBoost}
O.~Messai, F.~Hachouf, and Z.~A. Seghir, ``Adaboost neural network and
  cyclopean view for no-reference stereoscopic image quality assessment,''
  \emph{Signal Processing: Image Communication}, vol.~82, p. 115772, 2020.

\bibitem{Classification}
A.~Krizhevsky, I.~Sutskever, and G.~E. Hinton, ``Imagenet classification with
  deep convolutional neural networks,'' \emph{Commun. ACM}, vol.~60, no.~6, p.
  84–90, may 2017.

\bibitem{segmentation}
J.~Long, E.~Shelhamer, and T.~Darrell, ``Fully convolutional networks for
  semantic segmentation,'' in \emph{2015 IEEE Conference on Computer Vision and
  Pattern Recognition (CVPR)}, 2015, pp. 3431--3440.

\bibitem{Detection}
S.~Ren, K.~He, R.~Girshick, and J.~Sun, ``Faster r-cnn: Towards real-time
  object detection with region proposal networks,'' in \emph{Advances in Neural
  Information Processing Systems 28 (NIPS 2015)}, ser. Advances in Neural
  Information Processing Systems, vol.~28, 2015.

\bibitem{Zhang2016}
W.~Zhang, C.~Qu, L.~Ma, J.~Guan, and R.~Huang, ``Learning structure of
  stereoscopic image for no-reference quality assessment with convolutional
  neural network,'' \emph{Pattern Recognition}, vol.~59, pp. 176--187, 2016.

\bibitem{Fang2019Siamese}
Y.~Fang, J.~Yan, X.~Liu, and J.~Wang, ``Stereoscopic image quality assessment
  by deep convolutional neural network,'' \emph{Journal of Visual Communication
  and Image Representation}, vol.~58, pp. 400--406, 2019.

\bibitem{Shi2020PR}
Y.~Shi, W.~Guo, Y.~Niu, and J.~Zhan, ``No-reference stereoscopic image quality
  assessment using a multi-task cnn and registered distortion representation,''
  \emph{Pattern Recognition}, vol. 100, p. 107168, 2020.

\bibitem{lsm-zhaoping}
S.~Li, P.~Zhao, and Y.~Chang, ``No-reference stereoscopic image quality
  assessment based on visual attention mechanism,'' in \emph{2020 IEEE
  International Conference on Visual Communications and Image Processing
  (VCIP)}, 2020, pp. 326--329.

\bibitem{Zhouwei2019TIP}
W.~Zhou, Z.~Chen, and W.~Li, ``Dual-stream interactive networks for
  no-reference stereoscopic image quality assessment,'' \emph{IEEE Transactions
  on Image Processing}, vol.~28, no.~8, pp. 3946--3958, 2019.

\bibitem{Bourbia2021ICIP}
S.~Bourbia, A.~Karine, A.~Chetouani, and M.~E. Hassoun, ``A multi-task
  convolutional neural network for blind stereoscopic image quality assessment
  using naturalness analysis,'' in \emph{2021 IEEE International Conference on
  Image Processing (ICIP)}, 2021, pp. 1434--1438.

\bibitem{Yan2020}
J.~Yan, Y.~Fang, L.~Huang, X.~Min, Y.~Yao, and G.~Zhai, ``Blind stereoscopic
  image quality assessment by deep neural network of multi-level feature
  fusion,'' in \emph{2020 IEEE International Conference on Multimedia and Expo
  (ICME)}, 2020, pp. 1--6.

\bibitem{1993topdown}
S.~M. Kosslyn, N.~M. Alpert, W.~L. Thompson, V.~Maljkovic, S.~B. Weise, C.~F.
  Chabris, S.~E. Hamilton, S.~L. Rauch, and F.~S. Buonanno, ``{Visual Mental
  Imagery Activates Topographically Organized Visual Cortex: PET
  Investigations},'' \emph{Journal of Cognitive Neuroscience}, vol.~5, no.~3,
  pp. 263--287, 07 1993.

\bibitem{2003topdown}
M.~Bar, ``{A Cortical Mechanism for Triggering Top-Down Facilitation in Visual
  Object Recognition},'' \emph{Journal of Cognitive Neuroscience}, vol.~15,
  no.~4, pp. 600--609, 05 2003.

\bibitem{shi2023topdown}
B.~Shi, T.~Darrell, and X.~Wang, ``Top-down visual attention from analysis by
  synthesis,'' in \emph{2023 IEEE/CVF Conference on Computer Vision and Pattern
  Recognition (CVPR)}, 2023, pp. 2102--2112.

\bibitem{TPAMI2013}
N.~Kruger, P.~Janssen, S.~Kalkan, M.~Lappe, A.~Leonardis, J.~Piater, A.~J.
  Rodriguez-Sanchez, and L.~Wiskott, ``Deep hierarchies in the primate visual
  cortex: What can we learn for computer vision?'' \emph{IEEE Transactions on
  Pattern Analysis and Machine Intelligence}, vol.~35, no.~8, pp. 1847--1871,
  2013.

\bibitem{ZhouMingyue}
M.~Zhou and S.~Li, ``Deformable convolution based no-reference stereoscopic
  image quality assessment considering visual feedback mechanism,'' in
  \emph{2021 International Conference on Visual Communications and Image
  Processing (VCIP)}, 2021, pp. 01--05.

\bibitem{Mitchell-iScience}
B.~Mitchell, K.~Dougherty, J.~Westerberg, B.~Carlson, L.~Daumail, A.~Maier, and
  M.~Cox, ``Stimulating both eyes with matching stimuli enhances v1
  responses,'' \emph{iScience}, vol.~25, p. 104182, 04 2022.

\bibitem{PKU2022}
\BIBentryALTinterwordspacing
S.-H. Zhang, X.~Zhao, S.-M. Tang, and C.~Yu, ``Neuronal responses to monocular
  and binocular stimulations in macaque v1 studied with two-photon calcium
  imaging,'' 09 2022. [Online]. Available:
  \url{https://www.researchgate.net/publication/363631428_Neuronal_responses_to_monocular_and_binocular_stimulations_in_macaque_V1_studied_with_two-photon_calcium_imaging}
\BIBentrySTDinterwordspacing

\bibitem{Shen2021}
L.~Shen, X.~Chen, Z.~Pan, K.~Fan, F.~Li, and J.~Lei, ``No-reference
  stereoscopic image quality assessment based on global and local content
  characteristics,'' \emph{Neurocomputing}, vol. 424, pp. 132--142, 2021.

\bibitem{Messai2022ICIP}
O.~Messai and A.~Chetouani, ``End-to-end deep multi-score model for
  no-reference stereoscopic image quality assessment,'' in \emph{2022 IEEE
  International Conference on Image Processing (ICIP)}, 2022, pp. 2721--2725.

\bibitem{Sim2022}
K.~Sim, J.~Yang, W.~Lu, and X.~Gao, ``Blind stereoscopic image quality
  evaluator based on binocular semantic and quality channels,'' \emph{IEEE
  Transactions on Multimedia}, vol.~24, pp. 1389--1398, 2022.

\bibitem{Si2022TIP}
J.~Si, B.~Huang, H.~Yang, W.~Lin, and Z.~Pan, ``A no-reference stereoscopic
  image quality assessment network based on binocular interaction and fusion
  mechanisms,'' \emph{IEEE Transactions on Image Processing}, vol.~31, pp.
  3066--3080, 2022.

\bibitem{SE}
J.~Hu, L.~Shen, and G.~Sun, ``Squeeze-and-excitation networks,'' in \emph{2018
  IEEE/CVF Conference on Computer Vision and Pattern Recognition}, 2018, pp.
  7132--7141.

\bibitem{CBAM}
S.~Woo, J.~Park, J.-Y. Lee, and I.~S. Kweon, ``Cbam: Convolutional block
  attention module,'' in \emph{Computer Vision -- ECCV 2018}, V.~Ferrari,
  M.~Hebert, C.~Sminchisescu, and Y.~Weiss, Eds.\hskip 1em plus 0.5em minus
  0.4em\relax Cham: Springer International Publishing, 2018, pp. 3--19.

\bibitem{non-local}
X.~Wang, R.~Girshick, A.~Gupta, and K.~He, ``Non-local neural networks,'' in
  \emph{2018 IEEE/CVF Conference on Computer Vision and Pattern Recognition},
  2018, pp. 7794--7803.

\bibitem{GC}
Y.~Cao, J.~Xu, S.~Lin, F.~Wei, and H.~Hu, ``Gcnet: Non-local networks meet
  squeeze-excitation networks and beyond,'' in \emph{2019 IEEE/CVF
  International Conference on Computer Vision Workshop (ICCVW)}, 2019, pp.
  1971--1980.

\bibitem{SK}
X.~Li, W.~Wang, X.~Hu, and J.~Yang, ``Selective kernel networks,'' in
  \emph{2019 IEEE/CVF Conference on Computer Vision and Pattern Recognition
  (CVPR)}, 2019, pp. 510--519.

\bibitem{resblock}
K.~He, X.~Zhang, S.~Ren, and J.~Sun, ``Deep residual learning for image
  recognition,'' in \emph{2016 IEEE Conference on Computer Vision and Pattern
  Recognition (CVPR)}, 2016, pp. 770--778.

\bibitem{resblock-v2}
------, ``Identity mappings in deep residual networks,'' in \emph{Computer
  Vision -- ECCV 2016}, B.~Leibe, J.~Matas, N.~Sebe, and M.~Welling, Eds.\hskip
  1em plus 0.5em minus 0.4em\relax Cham: Springer International Publishing,
  2016, pp. 630--645.

\bibitem{ConnecAttn}
X.~Ma, J.~Guo, S.~Tang, Z.~Qiao, Q.~Chen, Q.~Yang, S.~Fu, P.~Palacharla,
  N.~Wang, and X.~Wang, ``Learning connected attentions for convolutional
  neural networks,'' in \emph{2021 IEEE International Conference on Multimedia
  and Expo (ICME)}, 2021, pp. 1--6.

\bibitem{Liu2020NC}
Y.~Liu, C.~Tang, Z.~Zheng, and L.~Lin, ``No-reference stereoscopic image
  quality evaluator with segmented monocular features and perceptual binocular
  features,'' \emph{Neurocomputing}, vol. 405, pp. 126--137, 2020.

\bibitem{dropout}
G.~E. Hinton, N.~Srivastava, A.~Krizhevsky, I.~Sutskever, and R.~R.
  Salakhutdinov, ``Improving neural networks by preventing co-adaptation of
  feature detectors,'' 2012.

\bibitem{LIVEI}
A.~K. Moorthy, C.-C. Su, A.~Mittal, and A.~C. Bovik, ``Subjective evaluation of
  stereoscopic image quality,'' \emph{Signal Processing: Image Communication},
  vol.~28, no.~8, pp. 870--883, 2013.

\bibitem{LIVEII-FR}
M.-J. Chen, C.-C. Su, D.-K. Kwon, L.~K. Cormack, and A.~C. Bovik,
  ``Full-reference quality assessment of stereopairs accounting for rivalry,''
  \emph{Signal Processing: Image Communication}, vol.~28, no.~9, pp.
  1143--1155, 2013.

\bibitem{LIVEII-NR}
M.-J. Chen, L.~K. Cormack, and A.~C. Bovik, ``No-reference quality assessment
  of natural stereopairs,'' \emph{IEEE Transactions on Image Processing},
  vol.~22, no.~9, pp. 3379--3391, 2013.

\bibitem{WIVCI}
J.~Wang and W.~Zhou, ``Perceptual quality of asymmetrically distorted
  stereoscopic images: The role of image distortion types,'' in
  \emph{International Workshop on Video Processing \& Quality Metrics for
  Consumer Electronics}, 2014.

\bibitem{WIVCII}
J.~Wang, A.~Rehman, K.~Zeng, S.~Wang, and Z.~Wang, ``Quality prediction of
  asymmetrically distorted stereoscopic 3d images,'' \emph{IEEE Transactions on
  Image Processing}, vol.~24, no.~11, pp. 3400--3414, 2015.

\bibitem{IQA-non-linear}
H.~Sheikh, M.~Sabir, and A.~Bovik, ``A statistical evaluation of recent full
  reference image quality assessment algorithms,'' \emph{IEEE Transactions on
  Image Processing}, vol.~15, no.~11, pp. 3440--3451, 2006.

\bibitem{Adam}
\BIBentryALTinterwordspacing
D.~P. Kingma and J.~Ba, ``Adam: A method for stochastic optimization,'' in
  \emph{ICLR (Poster)}, 2015. [Online]. Available:
  \url{http://arxiv.org/abs/1412.6980}
\BIBentrySTDinterwordspacing

\bibitem{loshchilov2017sgdr}
\BIBentryALTinterwordspacing
I.~Loshchilov and F.~Hutter, ``{SGDR}: Stochastic gradient descent with warm
  restarts,'' in \emph{International Conference on Learning Representations},
  2017. [Online]. Available: \url{https://openreview.net/forum?id=Skq89Scxx}
\BIBentrySTDinterwordspacing

\bibitem{YangJiachen-IoT}
J.~Yang, B.~Jiang, H.~Song, X.~Yang, W.~Lu, and H.~Liu, ``No-reference
  stereoimage quality assessment for multimedia analysis towards
  internet-of-things,'' \emph{IEEE Access}, vol.~6, pp. 7631--7640, 2018.

\bibitem{Sun2020TMM}
G.~Sun, B.~Shi, X.~Chen, A.~S. Krylov, and Y.~Ding, ``Learning local
  quality-aware structures of salient regions for stereoscopic images via deep
  neural networks,'' \emph{IEEE Transactions on Multimedia}, vol.~22, no.~11,
  pp. 2938--2949, 2020.

\bibitem{YangJiachen-auto}
J.~Yang, K.~Sim, X.~Gao, W.~Lu, Q.~Meng, and B.~Li, ``A blind stereoscopic
  image quality evaluator with segmented stacked autoencoders considering the
  whole visual perception route,'' \emph{IEEE Transactions on Image
  Processing}, vol.~28, no.~3, pp. 1314--1328, 2019.

\end{thebibliography}


\begin{IEEEbiography}[{\includegraphics[width=1in,height=1.25in,clip,keepaspectratio]{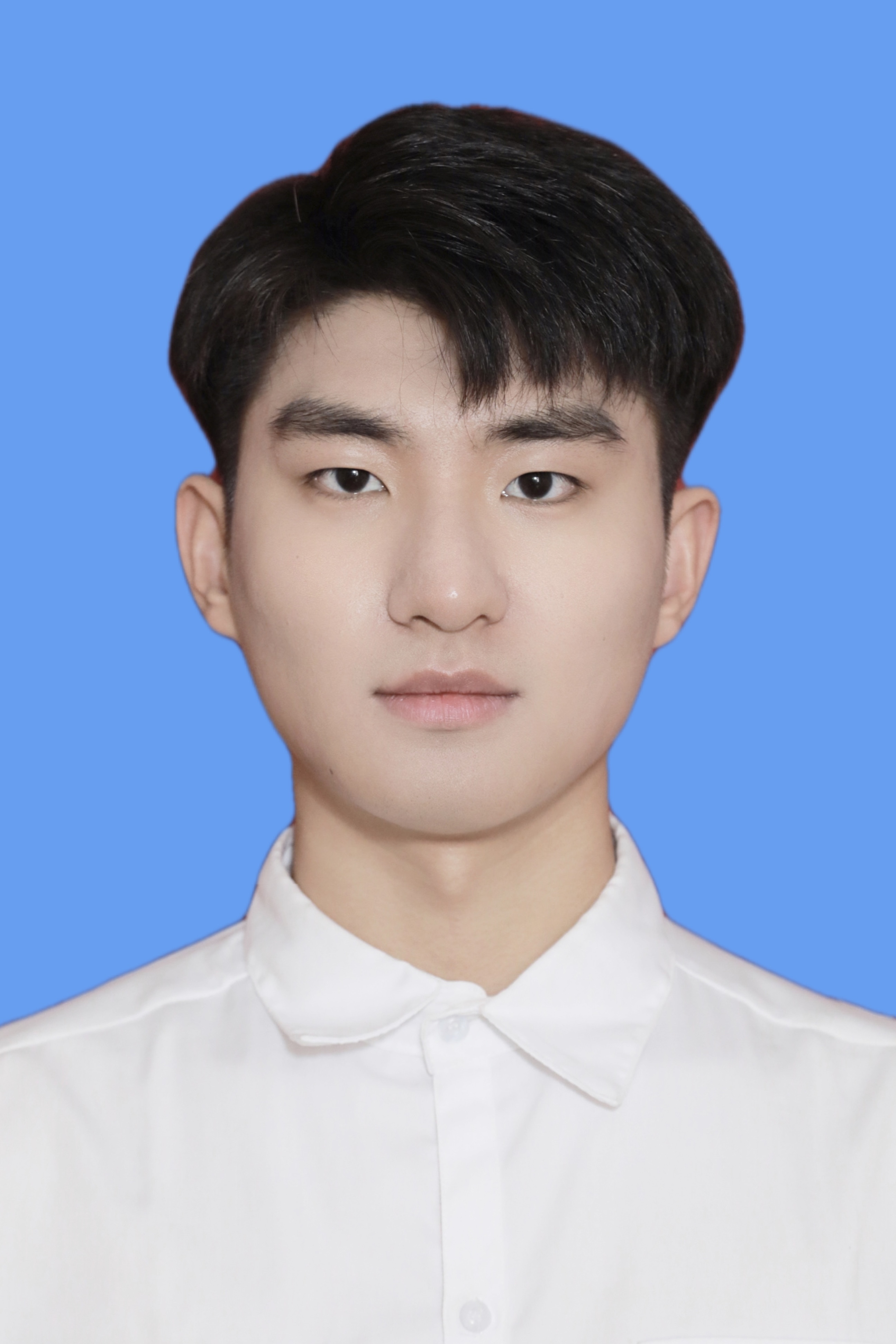}}]{Huilin Zhang}
received his Bachelor's degree from Tianjin University, Tianjin, China in 2021. He is currently pursuing a Master's degree at the School of Electrical and Information Engineering, Tianjin University. His research interests include 3D image processing and quality evaluation, attention mechanisms, neural networks, and deep learning.
\end{IEEEbiography}

\begin{IEEEbiography}[{\includegraphics[width=1in,height=1.25in,clip,keepaspectratio]{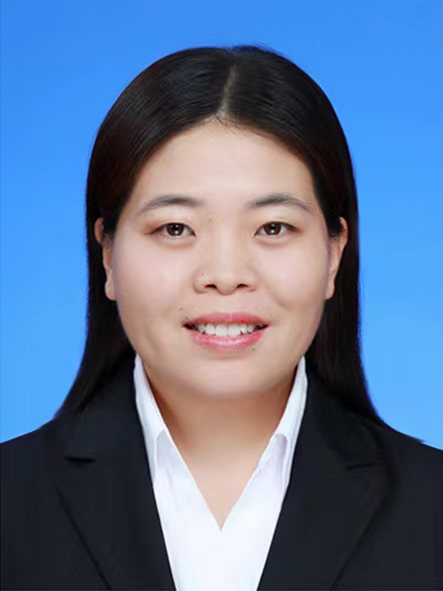}}]{Sumei Li}
received the Ph.D. from Nankai University, Tianjin, China. Since 2006, she has been an Associate Professor with the Communication Engineering Department, Tianjin University. She chaired a national 863 project, a National Natural Science Foundation, and a key fund in Tianjin. Her research interests include 3D image/video transmission, processing and quality evaluation, depth/image SR reconstruction, sparse representation, neural networks, and deep learning. Dr. Li is a member of China’s neural network committee.
\end{IEEEbiography}

\begin{IEEEbiography}[{\includegraphics[width=1in,height=1.25in,clip,keepaspectratio]{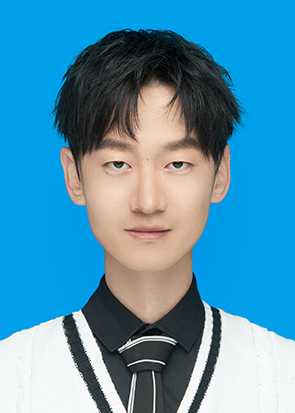}}]{Haoxiang Chang}
received the Bachelor's degree from Tianjin University, Tianjin, China, in 2022. He is currently working toward the Master's degree at the School of Electrical and Information Engineering, Tianjin University. His research interests include 2D/3D image quality evaluation, neural networks, and deep learning.
\end{IEEEbiography}

\begin{IEEEbiography}[{\includegraphics[width=1in,height=1.25in,clip,keepaspectratio]{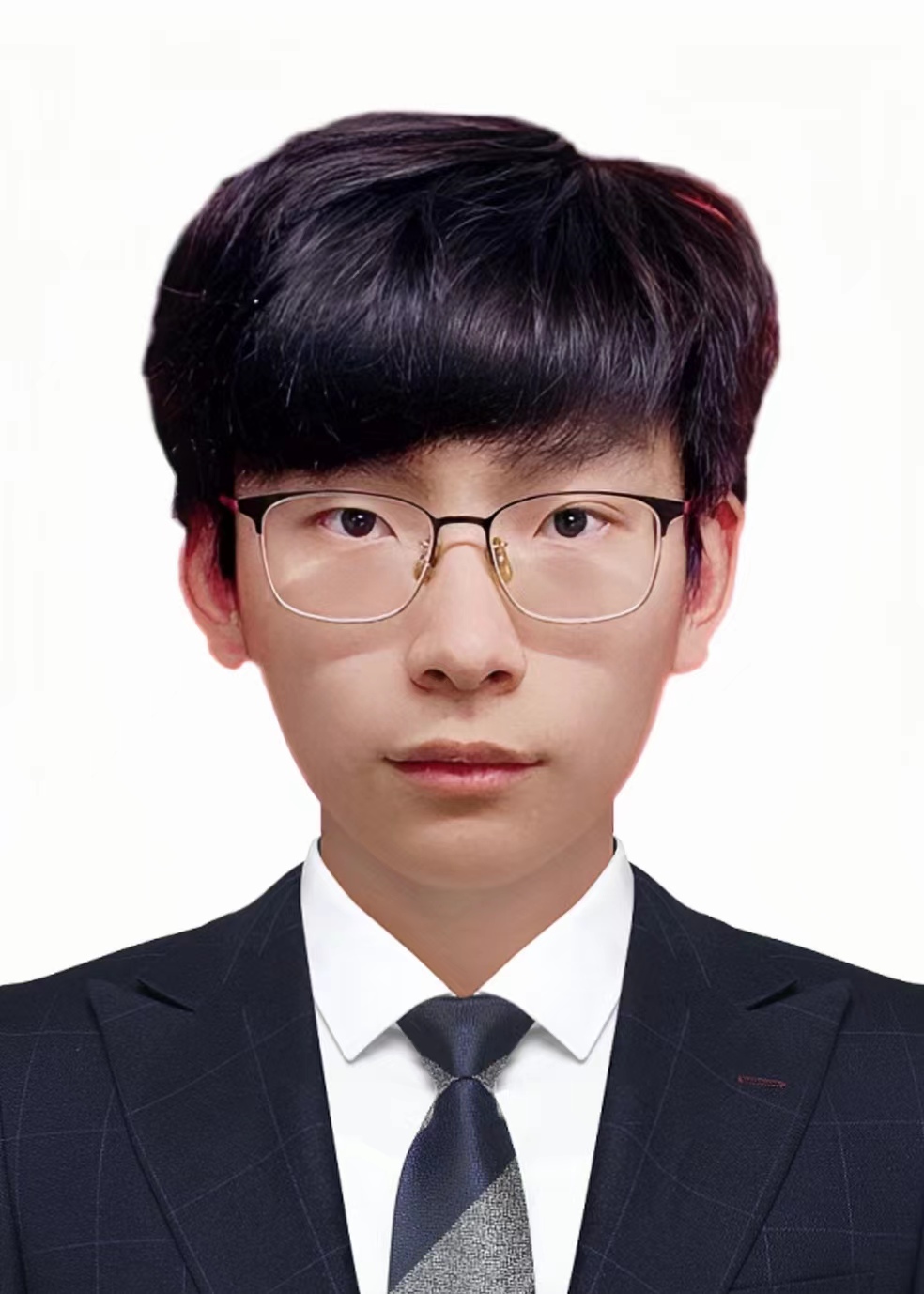}}]{Peiming Lin}
received the Bachelor's degree from Tianjin University, Tianjin, China, in 2022. He is currently working toward the Master's degree at the School of Electrical and Information Engineering, Tianjin University. His research interests include super-resolution, 3D image quality evaluation, and deep learning.
\end{IEEEbiography}


\vfill

\end{document}